\pdfoutput=1

\PassOptionsToPackage{table}{xcolor}

\documentclass{article}

\usepackage{arxiv}
\usepackage[utf8]{inputenc} 
\usepackage[T1]{fontenc}    

\usepackage{bm}
\usepackage{amsfonts}
\usepackage{amsmath}

\usepackage{booktabs}
\usepackage{tabularx}
\usepackage{hhline}
\usepackage{xspace}
\usepackage[table]{xcolor}
\usepackage{multirow}
\usepackage{subcaption}
\usepackage{array}
\newcolumntype{L}[1]{>{\raggedright\let\newline\\\arraybackslash\hspace{0pt}}m{#1}}
\newcolumntype{C}[1]{>{\centering\let\newline\\\arraybackslash\hspace{0pt}}m{#1}}
\newcolumntype{R}[1]{>{\raggedleft\let\newline\\\arraybackslash\hspace{0pt}}m{#1}}

\usepackage{perpage}
\MakePerPage{footnote}

\usepackage{amssymb}
\usepackage{pifont}
\newcommand{\cmark}{\ding{51}}%
\usepackage{graphicx}

\newcommand{\bigo}{\mathcal{O}}

\usepackage{dsfont}

\title{ProGraML: Graph-based Deep Learning for\\Program Optimization and Analysis}

\author{%
	Chris Cummins\thanks{Both authors contributed equally}\\
	School of Informatics \\
	University of Edinburgh \\
	\texttt{c.cummins@ed.ac.uk} \\
	\And
	Zacharias V. Fisches\textsuperscript{*}\\
	Department of Computer Science \\
	ETH Zurich \\
	\texttt{zfisches@student.ethz.ch} \\
	\And
	Tal Ben-Nun \\
	Department of Computer Science \\
	ETH Zurich \\
	\texttt{talbn@inf.ethz.ch} \\
	\And
	Torsten Hoefler \\
	Department of Computer Science \\
	ETH Zurich \\
	\texttt{htor@inf.ethz.ch} \\
	\And
	Hugh Leather \\
	School of Informatics \\
	University of Edinburgh \\
	\texttt{hleather@inf.ed.ac.uk} \\
}

\begin{document}

\maketitle
\begin{abstract}
The increasing complexity of computing systems places a tremendous burden on optimizing compilers, requiring ever more accurate and aggressive optimizations. Machine learning offers significant benefits for constructing optimization heuristics but there remains a gap between what state-of-the-art methods achieve and the performance of an optimal heuristic. Closing this gap requires improvements in two key areas: a \emph{representation} that accurately captures the semantics of programs, and a \emph{model architecture} with sufficient expressiveness to reason about this representation.

We introduce \textsc{ProGraML} --- \emph{Program Graphs for Machine Learning} --- a novel graph-based program representation using a low level, language agnostic, and portable format; and machine learning models capable of performing complex downstream tasks over these graphs. The \textsc{ProGraML} representation is a directed attributed multigraph that captures control, data, and call relations, and summarizes instruction and operand types and ordering. Message Passing Neural Networks propagate information through this structured representation, enabling whole-program, per-instruction, and per-variable classification tasks. \textsc{ProGraML} is a compiler-independent representation with support currently for LLVM and XLA IRs.

\textsc{ProGraML} provides a general-purpose program representation that equips learnable models to perform the types of program analysis that are fundamental to optimization. To this end, we evaluate the performance of our approach first on a suite of traditional compiler analysis tasks: control flow reachability, dominator trees, data dependencies, variable liveness, and common subexpression detection. On a benchmark dataset of 250k LLVM-IR files covering six source programming languages, \textsc{ProGraML} achieves an average 94.0 $F_1$ score, significantly outperforming the state-of-the-art approaches. We then apply our approach to two high-level tasks --- heterogeneous device mapping and program classification --- setting new state-of-the-art performance in both.
\end{abstract}

\section{Introduction}

The landscape of computing ecosystems is becoming increasingly complex: multi-core and many-core processors, heterogeneous systems, distributed and cloud platforms. Manually extracting performance and energy benefits from systems like these is beyond the capabilities of most programmers. In such an environment, high quality optimization heuristics are not just desirable, they are required. Despite this, good optimization heuristics are hard to come by.

\begin{figure}
	\centering %
	\includegraphics[width=.65\linewidth]{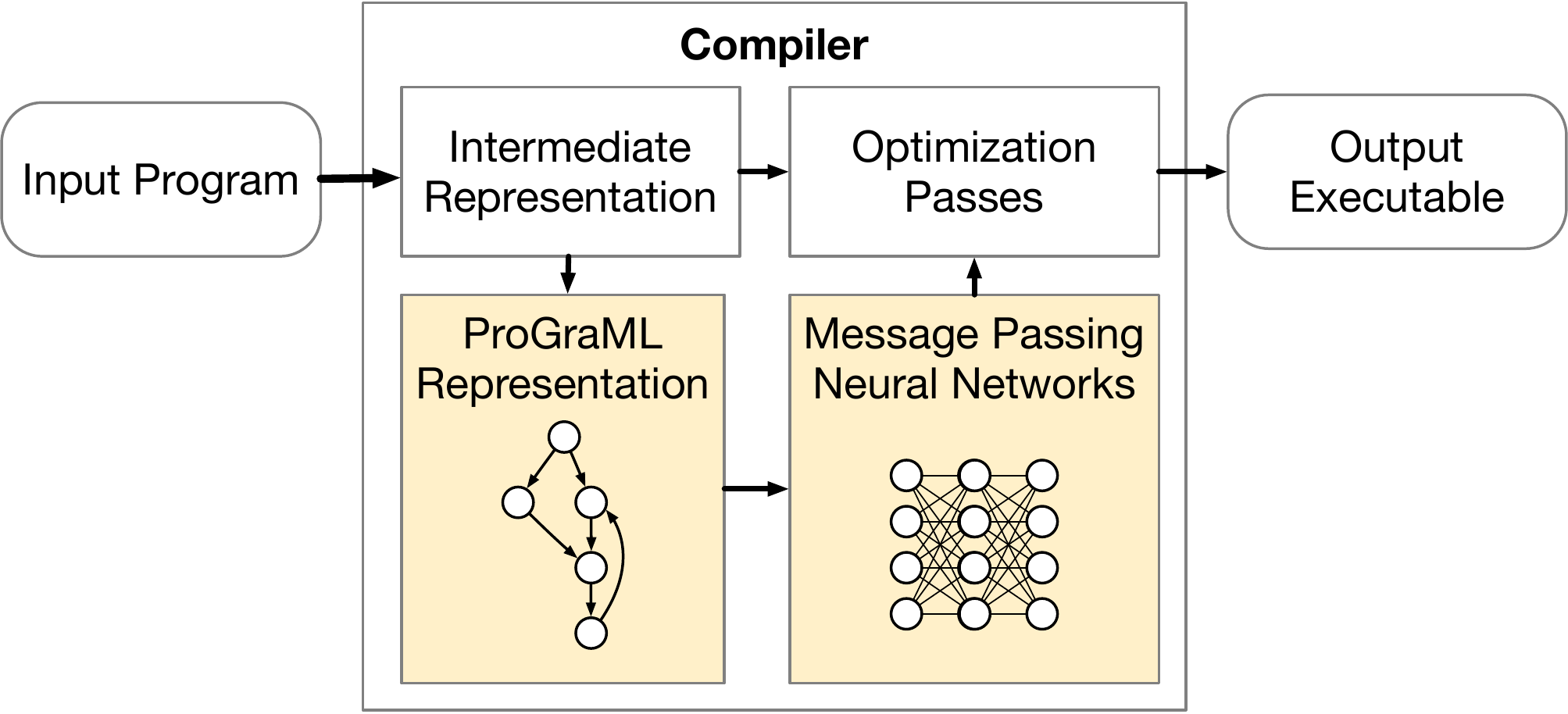}
	\caption{Our proposed approach for compiler analyses driven by graph-based deep learning. The \textsc{ProGraML} representation is derived from a compiler's IR and serves as input to Message Passing Neural Networks, which provide optimization decisions in place of traditional handwritten heuristics.}%
	\label{figure:overview}%
\end{figure}

Designing and tuning optimization heuristics takes time, effort, and resources. To make things worse, this is a Sisyphean task: even minor changes in a development toolchain might require retuning the heuristic; major changes in the software or the hardware usually require freshly designed heuristics. Machine learning offers to liberate us from this cycle by replacing fragile hand-tuned optimization heuristics with models that are inferred automatically from real performance data~\cite{Ashouri2018,Wang2018}. Typically, programs are represented using a sequence of numerical features which are derived from programs using ad-hoc analysis, but such approaches fail to capture the rich semantic structure of programs, limiting the ability for models to reason about program behavior. This is easy to see in traditional machine learned approaches, where the designer explicitly chooses a program representation based only on a few properties deemed important.  Such representations prevent models from reproducing the reasoning of even basic compiler analyses and that, in turn, limits their ability to make good optimization decisions.

Recent deep learning approaches that work directly on code~\cite{Allamanis2017a} are limited, both in the way they represent the inputs and in the way they process them. Representations based on source code and its direct artifacts (e.g. AST) put unnecessary emphasis on naming and stylistic choices that might or might not correlate with the functionality of the code~\cite{Alon2018a,Yin2018,Haj-Ali2019a}. Current IR-based approaches~\cite{Ben-nun2018,Mirhoseini2017,Brauckmann2020} use compilation to remove such noise but in the process they omit important information about the program.

In both cases, the model is expected to reason about the flow of information in the program using representations that do not encode this information clearly. If this was not difficult enough already, processing the code representations sequentially, as most existing approaches do, makes it practically impossible. Related statements can easily be separated by hundreds of lines of irrelevant code in sequential representations. Due to \textit{vanishing gradients}~\cite{Bengio1994} and \textit{catastrophic forgetting}~\cite{McCloskey1989}, neural networks are unlikely to even notice such very long range dependencies.

In this paper, we propose overcoming this limitation by making the program's control, data, and call dependencies a central part of the program's representation \emph{and} a primary consideration when processing it. We achieve this by seeing the program as a graph, in which individual statements are connected to other statements through relational dependencies. The latent representation of each statement is then a function of not just the statement itself but the latent representations of its graph neighborhood. Each statement and data element in the program is understood only in the context of the statements interacting with it. In contrast to prior sequential learning systems for code, this representation closely resembles the intermediate representations used by compilers, and the propagation of information through these graphs mimics the behavior of typical iterative data-flow analyses.

Techniques for learning over graphs have recently been proposed and have shown promise in a number of domains~\cite{Li2015a,Schlichtkrull2018a,Gilmer2017}. Our intent with this work is to extend these approaches to the domain of program analysis and provide a systematic evaluation of their suitability and limits for compiler tasks. With a graph-based approach, we are able to automatically learn established compiler analyses that rely on control and data flow and do it far better than existing code modeling approaches. Downstream tasks built on top of such graph models can then natively incorporate approximate compiler analyses into their decision making, leading to superior and more powerful models.

\newpage
\subsection{Contributions}

We make the following contributions:

\begin{itemize}
	\item A portable, language-agnostic graph representation of programs derived from compiler intermediate representations (IR), and machine learning models for relational reasoning about the control, data, and call relations in programs\footnote{Code and datasets available at \texttt{https://chriscummins.cc/ProGraML}}. The \textsc{ProGraML} representation is compiler-angostic, with support currently for LLVM and XLA IRs, and is the first to summarize instructions, operand types, and operand order.
	\item As a benchmark for our approach, we pose a suite of established compiler analysis tasks as supervised machine learning problems, comparing the performance of models using \textsc{ProGraML} against state-of-the-art code representations. Succeeding at these benchmark tasks requires the ability to model control- and data-flow, function boundaries, instruction types, and the type and order of operands. On a large dataset of over 250k LLVM-IRs taken from real-world software projects, our approach achieves an average 94.0 $F_1$ score (98.9 precision, 92.0 recall), a 3.44$\times$ improvement over the state-of-the-art.
	\item We demonstrate the efficacy of our approach on two challenging downstream compiler tasks: heterogeneous device mapping, and program classification. Our results in heterogeneous device mapping show the superiority of graph based approaches. In an ablation study in program classification, we show a significant performance improvement due both to the rich structure of the \textsc{ProGraML} representation and the message passing neural network approach. Our approach reaches an accuracy of 96.22\% in program classification, a $1.27\times$ improvement over the state-of-the-art. 
\end{itemize}

\section{Motivation}

Machine learning promises significant benefits for simplifying and improving the construction of optimization heuristics by replacing fragile and expensive hand-tuned heuristics with data-driven statistical modeling. For this goal to be realized, we require machine learning systems capable of reasoning about program semantics. Despite tremendous gains in recent years, state-of-the-art approaches for learning on code are not sufficiently powerful to replicate the analysis tasks that are fundamental to compilers. The crux of the problem lies in two central aspects of machine learning: input representation and model algorithmic complexity.

\paragraph{(I) Input Representation}

\begin{figure}[t]
	\centering
    \begin{subfigure}{.48\linewidth}
	    \includegraphics[width=\linewidth]{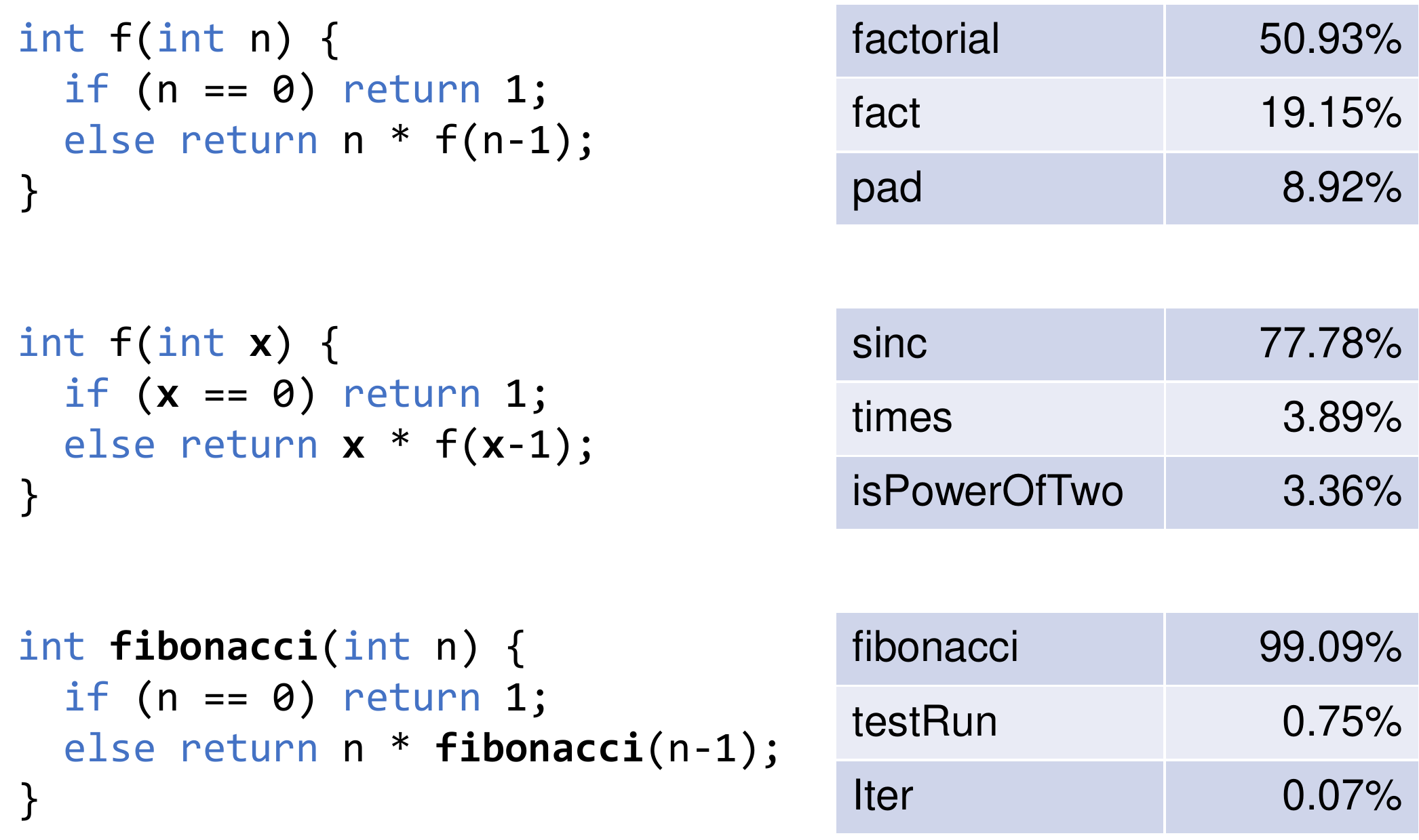}
	    \caption{code2vec~\cite{Alon2018a}.}
	    \label{subfig:code2vec}
	\end{subfigure}
	\hfill
	\begin{subfigure}{.29\linewidth}
		\vspace{2.5em}
	    \includegraphics[width=\linewidth]{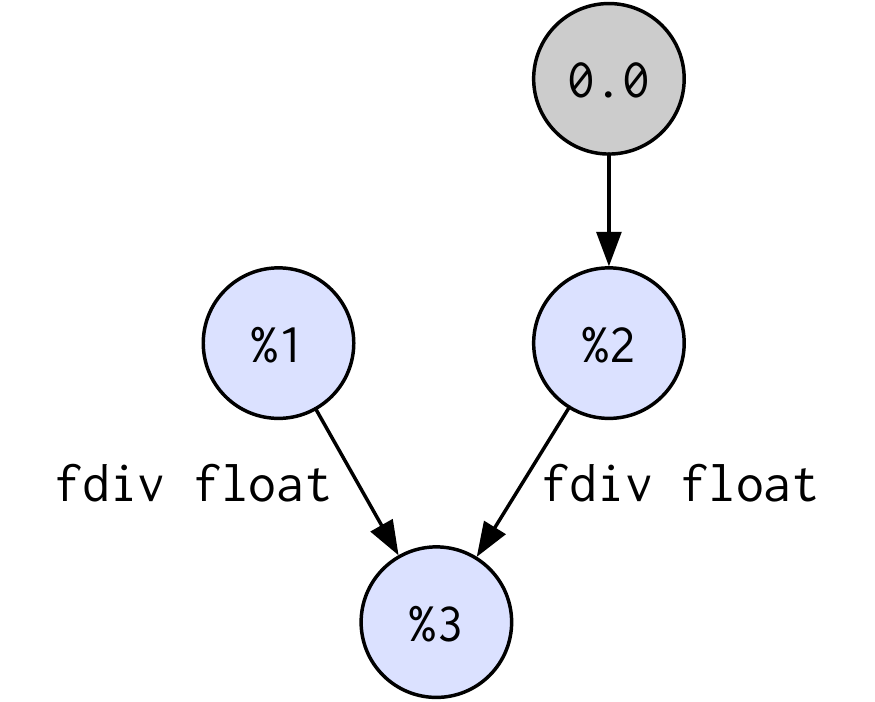}
	    \caption{XFG~\cite{Ben-nun2018}.}
	    \label{subfig:inst2vec}
	\end{subfigure}
	\hfill
	\begin{subfigure}{.18\linewidth}
	\vspace{2.4em}
	\includegraphics[width=\linewidth]{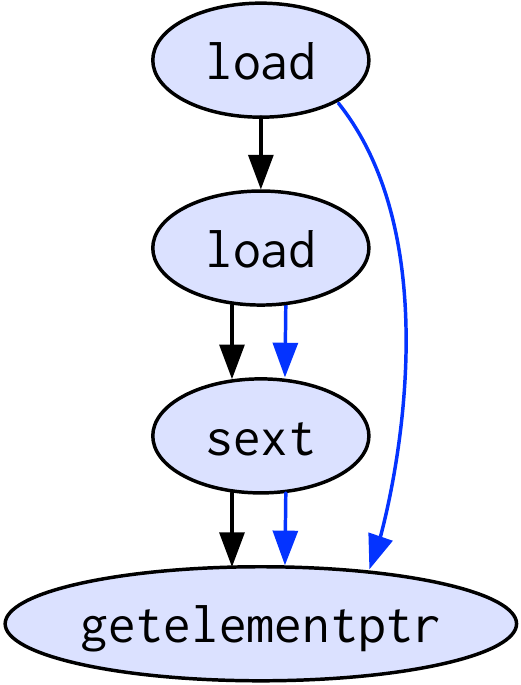}
	\caption{CDFG~\cite{Brauckmann2020}.}
	\label{subfig:cdfg}
	\end{subfigure}
	\caption{Limitations in state-of-the-art learnable code representations. In~(\subref{subfig:code2vec}) the model over-emphasizes identifier names such that the same algorithm produces three different classifications by changing the name of a function. The data-flow representation of~(\subref{subfig:inst2vec}) does not capture operand order, such that non-commutative statements such as division are indistinguishable. In~(\subref{subfig:cdfg}) control and data relations are captured, but both type information and operand order are omitted. Our approach is insensitive to identifier names and preserves operand order and type information.}
\end{figure}

To be processed by a Neural Network (NN), code inputs may be encoded either directly from a source language, by way of AST analysis, or using an IR. Examples of each abound in the literature~\cite{Allamanis2017a,Chen2019,Wang2018}. One state-of-the-art encoder, code2vec~\cite{Alon2018a}, uses AST paths to embed programs. code2vec proves highly effective at software engineering tasks such as algorithm classification, where the code was written by humans. However, as shown in Figure \ref{subfig:code2vec}, the trained representation can put more weight on names rather than code structure, where minute modifications completely change classification outcome. There are many benefits to such a representation, including smart pasting and automated refactoring. However, when analyzing code for optimization, identifier names are rarely of use, whereas structure and semantics should be the primary consideration. For example, in choosing to represent code at the AST level, the code2vec representation does not capture the data or control relations between statements.

An alternate approach which emphasizes semantics is Neural Code Comprehension~\cite{Ben-nun2018}, where an encoder uses Contextual Flow Graphs (XFG) built from LLVM-IR statements to create inputs for neural networks. The XFG combines partial data- and control-flow to represent the context of an individual statement. The statements are then mapped to latent-space representations using their neighbors in that graph. However, in partially combining DFGs and CFGs, the XFG representation omits important information such as order of instruction operands (as shown in Figure~\ref{subfig:inst2vec}), and the representation fails to capture execution order, critical for many optimization tasks.

A recent LLVM-IR representation uses Control and Data Flow Graphs (CDFG)~\cite{Brauckmann2020}. This representation makes the control and data relations between instructions explicit, but uses only instruction opcodes to compute latent representations. This omits information about programs that may be critical for optimization, such as data types, the presence of variables and constants, and the ordering of operands, shown in Figure~\ref{subfig:cdfg}.

Each of the three methods of encoding programs as inputs to neural networks omit information that is vital for compiler analysis. This hinders the ability of machine learning models to reason about optimizations and their impact on program behavior. To address these shortcomings we require an input representation that captures all parts of a program's semantics that are required for performing such analyses.

\paragraph{(II) Model Complexity}

The range of core operators in deep learning on code is largely confined to recurrent units (e.g. RNN, LSTM, GRU) on sequences. This poses limitations on the representation space of any such network's outputs. Take, for instance, dominator tree construction. An LSTM iterating forward over input code will not be able to solve the task \emph{by definition}, as statements needs to be analyzed backwards with respect to dependencies. Yet, the neural network only maintains $\bigo(1)$ memory capacity w.r.t. program length. Iterating in the opposite direction, in addition to being problem-specialized, would still not suffice, as dependencies may ``vanish'' in the corresponding gradients if dependent statements are far away from each other, such as in the presence of diverging control flow.

One way to increase the algorithmic complexity of the model is by allowing it to increase the number of code tokens that are processed simultaneously during inference. This approach, commonly used in literature by Transformer Networks~\cite{Vaswani2017}, use preceding tokens (unidirectional encoding) or preceding and subsequent tokens (bidirectional encoding) to learn \textit{attention matrices}. Such matrices ``focus'' the network on certain subsets of tokens, skipping others. However, this approach scales quadratically in memory and computation with the number of tokens.

Unlike in natural language text, the dependency structure of code is made explicit during compilation. We can thus employ domain-specific knowledge to construct the attention matrices in a scalable manner, using a graph representation of the tokens with dependencies as edges. A graph representation not only enables meaningful attention learning, but also facilitates propagating information across the graph in a manner similar to typical compiler analyses. Within the same step, a recurrent unit generates $\bigo(1)$ activations, whereas a graph NN generates $\bigo(|V|)$. 

To demonstrate this expressive power, let us consider control-flow reachability analysis as a learning task. The goal of the model is to tag statements that are reachable from one or more given tagged statements. With a sequential LSTM, the model would have to be trained to memorize nodes along some linear order of the given program. With an unbounded number of nodes to track and variably-sized regions to skip, the task becomes infeasible. A message-passing graph neural network, in contrast, needs only to learn to pass a message forward in the case of an existing control-flow edge between two nodes, essentially learning an identity operation over control-flow edges and zero on others.

In this work, we overcome the above limitations of prior model and representation approaches, leveraging the graph structure of IR code, taking path analysis and the semantics of individual statements into account. 
\section{A Graphical Program Representation for Analysis and Optimization}
\label{sec:graph-representation}

For machine learning to successfully reason over programs, a suitable input representation must be used. This section presents \textsc{ProGraML}, a novel program representation that closely matches the representations used traditionally within compilers and can be processed natively by machine learning models. Unlike prior approaches that rely on hand-engineered feature extractors~\cite{Ashouri2018,Wang2018} or which are closely tied to the syntax of the target program language~\cite{Allamanis2017b}, our approach captures whole-program control, data, and call relations and is both task- and language-agnostic.

\begin{figure*}
  \centering %
  \begin{subfigure}{.48\linewidth}%
  	\centering
  	\includegraphics[width=\linewidth]{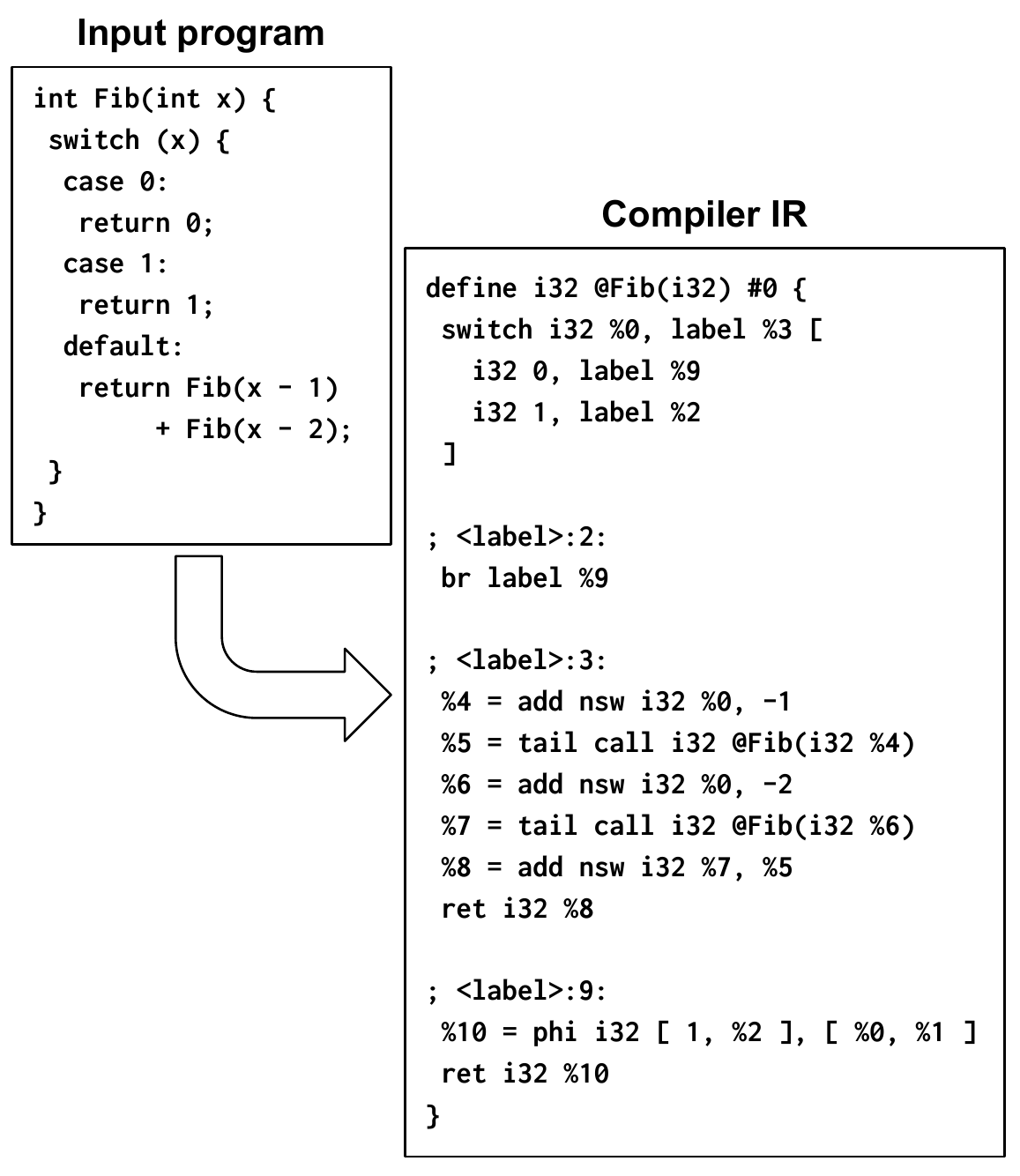}%
  	\caption{Compiler intermediate representation.}
  	\label{subfigure:ir}%
  \end{subfigure}
	\begin{subfigure}{.48\linewidth}%
		\centering
  	\includegraphics[width=\linewidth]{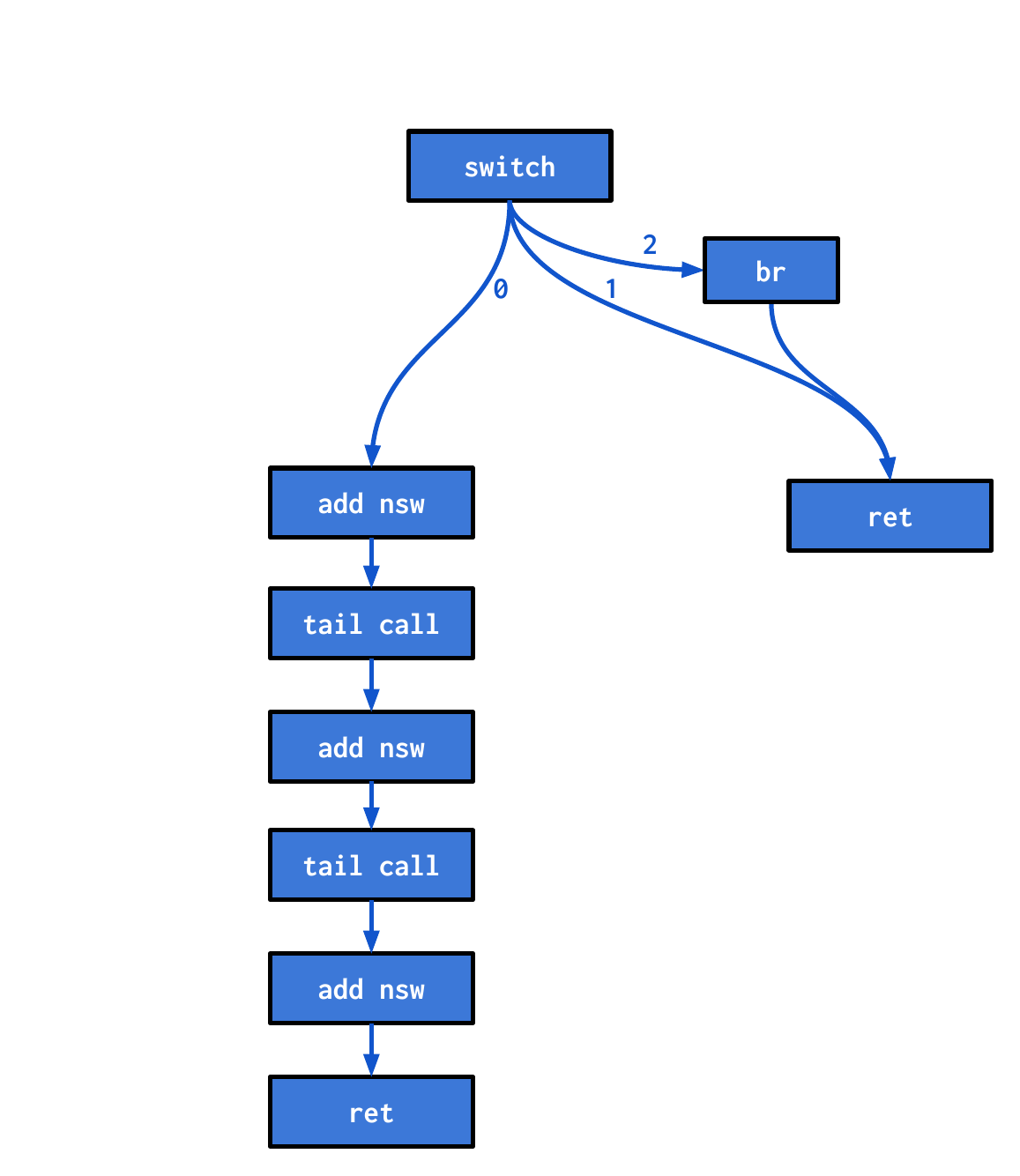}%
  	\caption{Construct control flow.}
  	\label{subfigure:control_flow}%
	\end{subfigure}
  \\*
  \vspace{1em}
  \begin{subfigure}{.48\linewidth}%
  	\includegraphics[width=\linewidth]{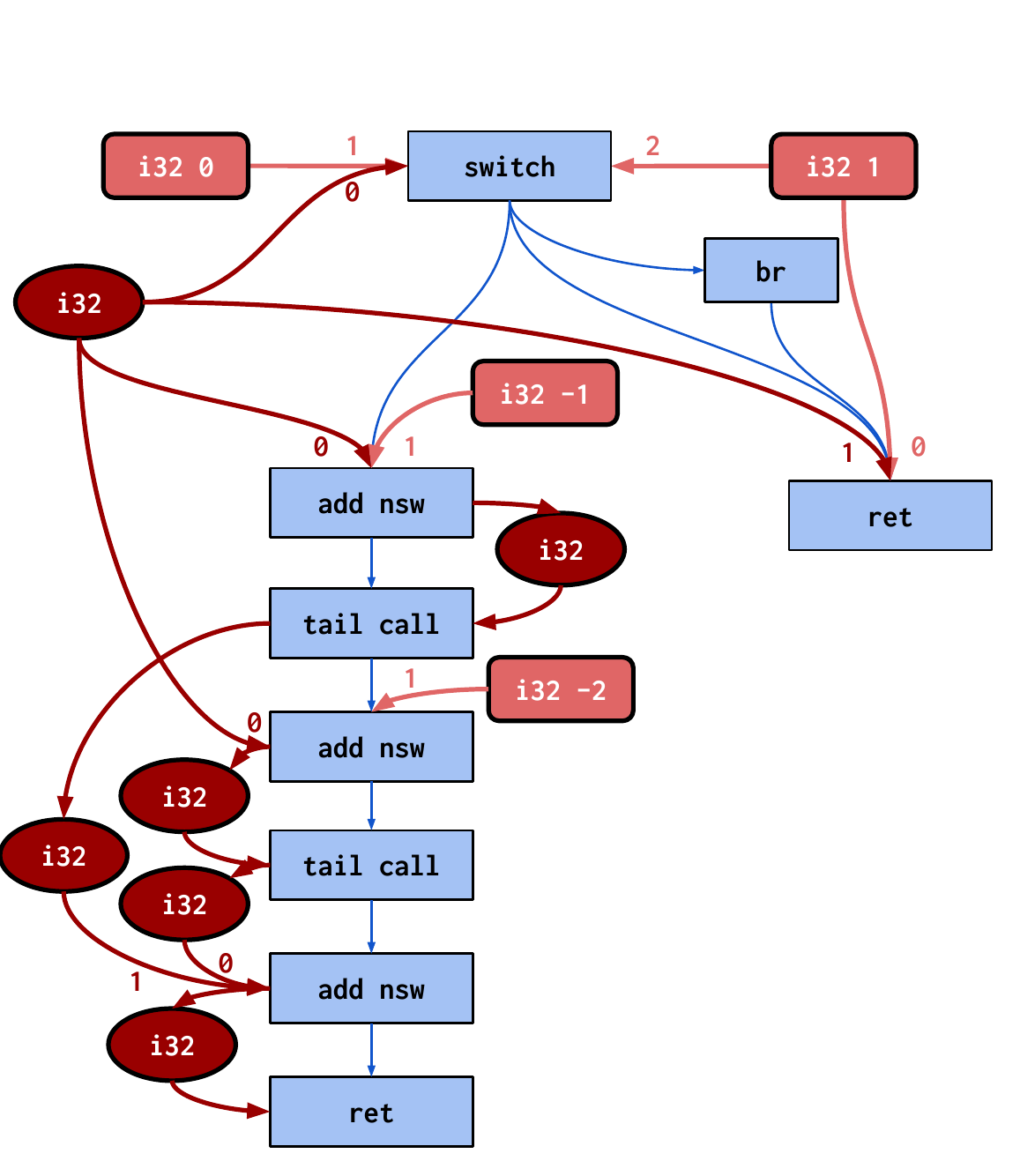}%
  	\caption{Add data flow for variables and constants.}
  	\label{subfigure:data_flow}%
	\end{subfigure}
  \begin{subfigure}{.48\linewidth}%
	  \includegraphics[width=\linewidth]{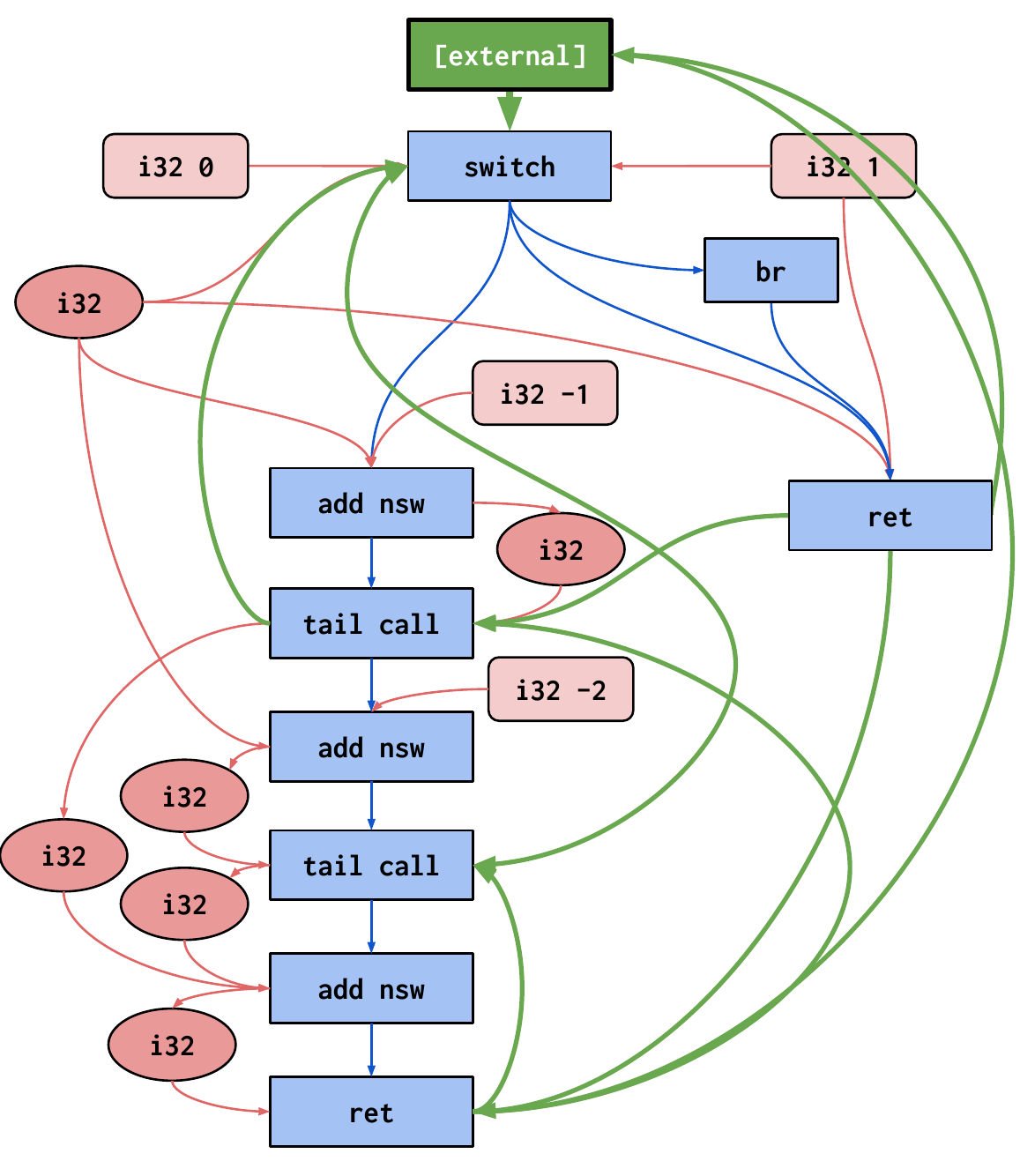}%
	  \caption{Add call flow for call sites.}
	  \label{subfigure:call_flow}%
	\end{subfigure}
  \caption{Construction of a \textsc{ProGraML} representation for a simple C implementation of recursive Fibonacci using LLVM-IR. The input program is passed through the compiler front end to produce an intermediate representation (a). A full flow graph is constructed from the IR statements and control flow edges inserted (b). Vertices for variables and constant values are  added and data-flow edges inserted (c). Finally, call sites are extracted and call edges inserted from call sites to function entry statements, and from function exit vertices to call sites (d). All edges are positional, for clarity we have omitted position labels where not required.}%
  \label{figure:graph_construction}%
\end{figure*}

\subsection{Overview}

The \textsc{ProGraML} representation of a compiler IR serves as the union between a call graph, control-flow graph, and data-flow graph. We represent programs as directed multigraphs where statements, identifiers, and immediate values are vertices, and relations between vertices are edges. Edges are typed to differentiate control-, data-, and call-flow. Additionally, we augment edges with a positional label to encode the order of operands for statements, and to differentiate between divergent branches in control-flow. The \textsc{ProGraML} representation is processed natively by our machine learning models, described in Section~\ref{sec:graph-based-machine-learning}.

\subsection{Graph Construction}

We construct a \textsc{ProGraML} graph $G = (V, E)$ by traversing a compiler IR. Graph construction is divided into three stages: control-flow, data-flow, and call-flow, though in practice the three stages can be combined in a single $\bigo{(|V|+|E|)}$ pass. The representation is compiler-agnostic, adding support for a new compiler requires only a parser for the IR. Currently we support LLVM~\cite{Lattner2004} and XLA HLO~\cite{Leary2017} IRs. Figure~\ref{figure:graph_construction} shows the graph construction approach.

\paragraph{(I) Control Flow} We construct a full-flow graph from an IR by inserting a graph vertex for each instruction and control flow edges between them, as shown in Figure~\ref{subfigure:control_flow}. All control edges are augmented with a numeric position label using an ascending sequence based on their order in the list of a vertex's outgoing control edges. For instructions with a single control successor, the position of the control edge is 0. For a branching statement with $n$ successor statements, the control edge positions are in the range $0 \le e_{\text{pos}} \le n$. We do not need to encode the source function or basic block~\cite{Lattner2004} of instructions as this information is captured implicitly in structure of the graph; basic blocks are regions of instructions with a single entry and exit control edge, and functions are disconnected subgraphs.

\paragraph{(II) Data Flow} We introduce additional graph vertices for constant values and variables, shown in Figure~\ref{subfigure:data_flow}. Data-flow edges are added to capture the relation from constants and variables to the instructions that use them as operands, and instructions to produced variables. Each unique variable and constant is a vertex, which implicitly models the scope of variables, and unlike the tokenized representations of prior machine learning works, variables in different scopes always map to distinct vertices and can thus be discerned. Similarly, constant values with the same textual representation in source code (such as the number \texttt{3.14} with \texttt{float} and \texttt{double} precision types) are distinguishable in our representation. As with control edges, data edges have a position label which encodes the order of operands for instructions. The latent representation of an IR statement is thus a function of the vertex representing the instruction and the vertices of any operand variables or constants, modulated by their order in the list of operands.

\paragraph{(III) Call Flow} Control edges do not span functions, such that an IR with functions $F$ produces $|F|$ disconnected subgraphs (the same is not true for data edges which may cross function boundaries, for example in the case of an global constant which is used within multiple functions of a program). Instead, the relation between a statement which calls a function and the called function is captured through call edges, shown in Figure~\ref{subfigure:call_flow}. An outgoing call edge is inserted from the calling statement to the entry statement of a function. Return call edges are added from all terminal statements of a function to the calling statement. Call edges do not use position labels as there is no ordering to be imposed between the call sites of a function. For IRs which support external linkage, an additional vertex is created representing an external callsite and connected to all externally visible functions. Similarly, if a call site references a function not defined in the current IR, a \emph{dummy} function definition is created consisting of a pair of entry and exit instruction vertices, and connected normally through call edges. A single dummy function definition is created for each externally defined function and shared across all call sites in the current IR.

\subsection{Comparison to Other Representations}

As an input for machine learning, what distinguishes \textsc{ProGraML} from prior works is its close proximity to the structured representations traditionally used within compilers for  program analysis and optimization. Specifically, it is distinct from prior machine learning representations in three key areas:
\begin{enumerate}
	\item as an IR-level representation, it is independent of the source language and accompanying variances such as code style and identifier naming that affect source-level representations~\cite{Alon2018a,Cummins2017a};
	\item by representing programs as graphs with explicit control, data, and call edge types \textsc{ProGraML} captures a greater range of intra-program relations than prior graph representations~\cite{Ben-nun2018,Allamanis2017b,Park2012};
	\item and in trading sequential for graph representations, we do not sacrifice local sequence order, such as the ordering of diverging control flow or the ordering of operands that is lost in prior representations~\cite{Ben-nun2018,Brauckmann2020}.
\end{enumerate}

Table~\ref{tab:representation_taxonomy} provides a comparison of \textsc{ProGraML} to several recent learned representations of code.

\begin{table}
	\centering
	\footnotesize
	\begin{tabular}{r L{3.6cm} L{2.5cm} L{1.8cm} L{1.8cm} L{1.8cm}}
		\toprule
		& Source Languages & Representation & Flow-sensitive? & Position-sensitive? & Value-sensitive? \\
		\midrule
		AST Paths~\cite{Alon2018c} & C\#, Java, JavaScript, Python & AST & &  & \cmark \\
		CDFG~\cite{Brauckmann2020} & OpenCL & IR Graph & \cmark &  &  \\
		code2vec~\cite{Alon2018a} & Java & AST & & & \cmark \\
		DeepTune~\cite{Cummins2017b} & OpenCL & Token Sequence &  & \cmark & \cmark \\
		DeepTune-IR\cite{Barchi2019a} & OpenCL & IR Token Sequence &  & \cmark & \\
		DeepTyper~\cite{Hellendoorn2018} & JavaScript & Token Sequence &  & \cmark & \cmark \\
		inst2vec~\cite{Ben-nun2018} & C++, OpenCL & IR Graph & \cmark & & \cmark \\
		\textbf{\textsc{ProGraML}} & \textbf{C, C++, Fortran, Haskell, OpenCL, Swift} & \textbf{IR Graph} & \cmark & \cmark & \cmark \\
		\bottomrule
	\end{tabular}
	\vspace{.3em}
	\caption{%
		Taxonomy of recent code representations for machine learning. We classify approaches based on the type of representation used and the sensitivity to three categories: \{control/data/call\}-flow, operand positions, and operand values. Prior approaches require a trade-off in representational power, e.g. substituting a position-sensitive token sequence for a flow-sensitive graph. \textsc{ProGraML} is the first representation to span all categories.%
	}
	\label{tab:representation_taxonomy}
\end{table}

\section{Graph-based Machine Learning}
\label{sec:graph-based-machine-learning}

We formulate our system in a Message Passing Neural Network (MPNN) framework~\cite{Gilmer2017,Li2015a} and implement a single unified model for all our experiments. Our design mimics the \emph{transfer functions} and \emph{meet operators} of classical iterative dataflow analysis~\cite{Kam1977,Cooper2003}, replacing the rule-based implementations with deep learning analogues (message and update functions). Those can be specialized through training to solve a diverse set of problems without human intervention or algorithm design.

\subsection{Overview}

We learn over \textsc{ProGraML} representations of compiler IRs by mapping graph vertices to an initial state vector using an embedding. The vertex states are updated iteratively in a sequence of message passing steps, where at each step a new vertex state is computed as a function of its previous state and the state of its neighboring vertices. Separate functions are learned to update vertex neighbors based on their relation type, be it control, data or call, and reverse edges enable backwards propagation of information. After repeating this process of updating vertex states for a fixed number of iterations a readout function is used to aggregate the vertex representations to a single graph-level vector or set of vertex-level vectors.

\subsection{Model Design}
The \textsc{ProGraML} model takes as input a directed graph with additional information as presented in Section~\ref{sec:graph-representation} and consists of three logical phases: input encoding, message propagation, and result readout.

\paragraph{(I) Input Encoding} Starting from the augmented graph representation $G = (V, E)$ introduced in Section \ref{sec:graph-representation}, we capture the semantics of the program graph vertices by mapping every instruction, constant, and variable vertex $v \in V$ to a vector representation $h_v^0 \in \mathbb{R}^{d}$ by lookup in a fixed size embedding table $C_
\text{IR} \in \mathbb{R}^{n \times d}$. The mapping from vertex to embedding vector $f: v \hookrightarrow h_v^0 \in C_\text{IR}$ must be defined for each IR, though the embeddings themselves can be learned during training.

\begin{figure}
	\centering
	\includegraphics[width=\linewidth]{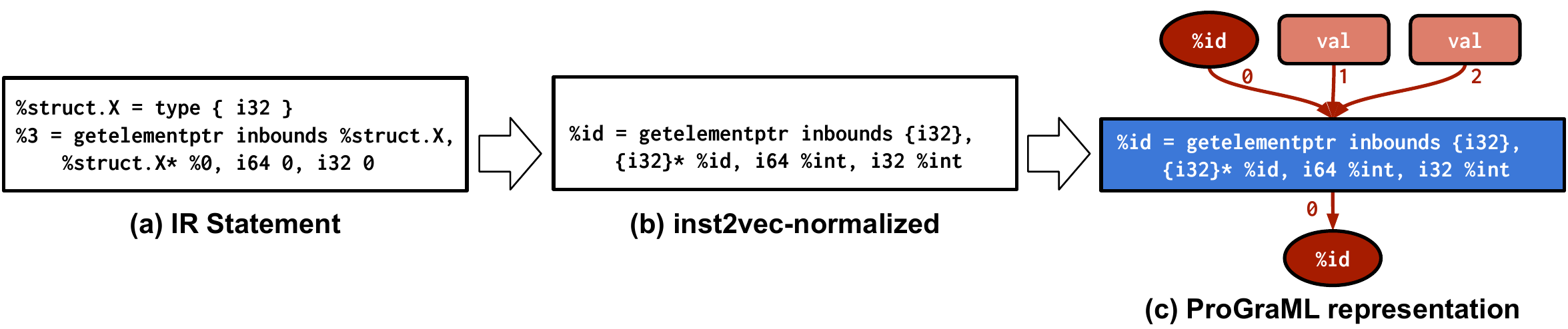}%
	\caption{Normalizing an LLVM-IR statement (a) by inlining type definitions and stripping identifiers using inst2vec~\cite{Ben-nun2018} (b). A normalized statement is used as the key for an embedding table lookups to produce the initial vector representation of a vertex. (c) shows the statement as contextualized in a \textsc{ProGraML} graph, where the operand variables and constants are data elements, differentiated by their position. Vertex labels represent embedding table keys. In this example, the statement is represented using five vertices and encoded with three unique embeddings.}%
	\label{fig:embedding}%
\end{figure}

For LLVM-IR we extend the inst2vec~\cite{Ben-nun2018} vocabulary, which represents 8,566 statements derived from a large corpus of LLVM-IR using 200-dimensional embeddings. Since the space of statements is unbounded, inst2vec uses a normalization process to inline type definitions and strip identifiers and immediate values, depicted in Figure~\ref{fig:embedding}. An inst2vec representation combines an instruction and its operands into a single token, so we augment the vocabulary with \textsc{Id} and \textsc{Val} tokens to represent variable and constant value vertices, respectively.

In expressing a statement as a combination of an instruction and its operands, our data-driven approach trades a certain amount of semantic resolution against good coverage of the vocabulary by the available datasets. The long tail of the distribution of statements jointly maps onto a special \textsc{Unknown} token vector in the vocabulary. In future work will simplify the LLVM-IR statement encoding by constructing separate vocabularies for instructions and operand types, increasing the expressiveness of the encoding by allowing a larger combination of instruction and operands to be represented.
Input \emph{vertex-selectors}, encoded as binary one-hot vectors, are used to mark the starting point for certain analyses and are concatenated to the vertex embeddings. 
Other global input features are used as auxiliary input features at readout time in step (III), where required.

\paragraph{(II) Message Propagation} Each iteration step is divided into a message propagation step followed by a vertex state update. During message propagation, each vertex in the graph collects learned messages $m_v^{t} \in \mathbb{R}^d$ from its undirected neighbors:

\begin{equation}
	m^t_{v} = \sum_{w \in \mathcal{N}(v)} A_{\mathrm{type}(e_{wv})}\,  (h_w^{t-1} \odot \textsc{pos}(e_{wv}))\,
\end{equation}

where $\odot$ denotes the Hadamard product and $e_{wv} \in E$ is the typed edge between vertex $w$ and $v$. In order to allow for reverse-propagation of messages, which is necessary for backward compiler analyses, we add backward edges for each edge in the graph. For backward edges we introduce separate parameters following Li et al.~\cite{Li2015a} to enable the network to distinguish between an edge and its backward sibling. During message propagation we scale the source states $h_v^{t-1}$ with \textsc{pos}$(\cdot) \in \mathbb{R}^d$, which is a constant sinusoidal position embedding~\cite{Vaswani2017,Gehring2017} that encodes the argument order of incoming (and outgoing) edges. This information is necessary for the network to distinguish non-commutative operations such as division.

The collected messages are subsequently used to update the vertex states $h_{v} \in \mathbb{R}^d$ in parallel according to an update function. In all our experiments, we employ a Gated Recurrent Unit (GRU) \cite{Cho2014} as our update function:

\begin{equation}
h_v^t = \textsc{Gru}(h_v^{t-1}, m_v^t)
\end{equation}

Step (II) is iterated $T$ times to extract vertex representations that are contextualized with respect to the given graph structure.

\paragraph{(III) Result Readout} We support whole program classification, per-statement classification, or per-identifier classification by employing different \emph{readout heads} on top of the iterated feature extraction: for graph-level classification we define a set function $R_G(\{h_v^T, h_v^0\}_{v \in V})$ that maps to class-scores, while for vertex-level inference, we separately map the extracted node features $h_v^T$ to probabilities $R_v(h_v^T, h_v^0)$ in parallel:
\begin{align}
	R_{v}(h_v^T, h_v^0) &= \sigma\left(i(h_v^T, h_v^0)\right) \cdot j(h_v^T) \\
	R_{G}(\{h_v^T, h_v^0\}_{v \in V}) &= \sum\limits_{v \in V}\,\,R_{v}(h_v^T, h_v^0)
\end{align}
where $i(\cdot)$ and $j(\cdot)$ are feed forward Neural Networks and $\sigma(\cdot)$ is the sigmoid activation function.
In the case where auxiliary graph-level features are available, those are concatenated to the readout values and fed through another feed forward Neural Network that employs Batch Normalization~\cite{Ioffe2015a} to allow for vastly different feature scales.

\section{Experimental Methodology}%
\label{sec:methodology}

We evaluate the effectiveness of our approach in three case studies. In the first, we apply our methodology to a suite of established compiler analysis tasks. These serve as demonstrations of the representational power of our approach and highlight the limitations both in prior machine learning approaches and in current MPNNs. The second case study then applies the approach to the challenging real-world optimization task of heterogeneous device mapping, comparing the performance of our model against state-of-the-art machine learning-based approaches. Finally, we apply our approach to the task of classifying algorithms from unlabelled implementations. This section describes the methodology used to construct these experiments: the datasets used, the model parameters, and training regimes.

\subsection{Case Study A: Compiler Analyses}

We construct a benchmark suite of traditional compiler analysis tasks to evaluate the representational power of our approach.  We chose a diverse set of tasks to capture a mixture of both forward and backward analyses, and control-, data-, and procedure-sensitive analyses. Our goal is not to suggest that machine learning should replace the existing implementations of these standard algorithms which can be found in any compiler, but rather, if a machine learning system is \emph{not} capable of producing these analyses, it stands to reason that performance on downstream tasks which depend on these analyses will suffer.

\subsubsection{Benchmark Analyses}

We selected five traditional compiler analyses to use as benchmarks for evaluating the representational power of our approach.

\paragraph{(I) Reachability} Control reachability is a fundamental compiler analysis which determines the set of statements that can be reached from a particular starting statement. Given $\text{succ}(n)$, which returns the control successors of statement $n$, the set of reachable statements starting at root $n$ can be found using forward analysis:

\begin{equation}
	\text{Reachable}(n) = \{n\} \cup \left( \bigcup_{s \in \text{succ}(n)} \text{Reachable}(s) \right)
\end{equation}

\paragraph{(II) Dominator trees} Statement $n$ dominates statement $m$ if every control flow path to $m$ passes through $n$. A dominator tree is the set of all nodes that dominate the statment at a particular program point. Like reachability, this analysis only requires propagation of control flow, but unlike reachability, dominator trees are typically constructed using backward analysis~\cite{Lengauer1979,Blazy2015}:

\begin{equation}
	\text{Dominators}(n) = \{n\} \cup \left( \bigcap_{p \in \text{pred}(n)} \text{Dominators}(p) \right)
\end{equation}

Where $\text{pred}(n)$ which returns the control predecessors of statement $n$.

\paragraph{(III) Live-out variables} A variable $a$ is live-out of statement $n$ if there exists some control successor of $n$ that uses $a$. Given $\text{uses}(n)$, which returns the operand variables of $n$, and $\text{defs}(n)$, which returns defined variables, the live-out variables can be computed forwards using:

\begin{equation}
	\text{LiveOut}(n) = \bigcup_{s \in \text{succ}(n)} \text{uses}(s) \cup \big(  \text{LiveOut}(s) - \text{defs}(s) \big)
\end{equation}

\paragraph{(IV) Data dependencies} The data dependencies of statement $n$ is the set of predecessor statements that must be evaluated to produce the operands of $n$. Computing data dependencies requires data sensitivity and is computed backwards:

\begin{equation}
	\text{DataDep}(n) = \text{defs}(n) \cup \left( \bigcup_{p \in \text{defs}(n)} \text{DataDep}(p) \right)
\end{equation}

Where $\text{defs}(n)$ returns the statements that produce operands of $n$.

\paragraph{(V) Global Common Subexpressions} The identification of common subexpressions is an important analysis for optimization. For compiler IRs we define a subexpression as a statement and its operands, ordered by either their position (for non-commutative operations), or lexicographically (for commutative operations). We thus formulate the common subexpression problem as, given a statement (which forms part of a subexpression), label any other statements in the program which compute the same subexpression. This is an inter-procedural analysis, though operands must obey their scope. Common subexpressions are typically identified using available expression analysis:

\begin{equation*}
\text{Avail}(n) = \text{uses}(n) \cup \left( \bigcap_{p \in \text{pred}(n)} \text{Avail}(p) \right) - \text{defs(n)}
\end{equation*}

Where $\text{uses}(n)$ return the expressions used by statement $n$, and $\text{defs}(n)$ returns the expressions defined by $n$.

\subsubsection{Datasets}

We assembled a large corpus of real-world LLVM-IRs from a variety of sources, summarized in Table~\ref{table:corpus}. We selected popular open source software projects that cover a diverse range of domains and disciplines, augmented by uncategorized code mined from popular GitHub projects using the methodology described by Cummins et al.~\cite{Cummins2017a}. Our corpus comprises a range of source languages (C, C++, Fortran, Haskell, OpenCL, and Swift) and exceeds 250k files. We de-duplicated the corpus first at the source level, then again after lowering to LLVM-IR. Lowering from source to IR was performed using the inst2vec methodology~\cite{Ben-nun2018}, in which an optimization level is chosen randomly per-file.

\begin{table}
	\centering%
\renewcommand{\arraystretch}{1.55}
\footnotesize
\begin{tabular}{L{3.35cm} L{1.45cm} L{2.75cm} | r r R{1.69cm} R{1.6cm}}
  & \textbf{Source language} & \textbf{Domain} & \textbf{IR files} & \textbf{IR lines} & \textbf{Avg. vertices / IR} & \textbf{Avg. edges / IR}\\
  \hline
  BLAS 3.8.0 & Fortran & Scientific Computing & 300 & 345,613 & 1,664 & 3,276\\
  \hline
  Linux 4.19 & C & Operating Systems & 13,851 & 41,332,089 & 1,857 & 3,760 \\
  \hline
  OpenCL Benchmarks~\cite{Cummins2017b} & OpenCL & Benchmarks & 256 & 149,779 & 1,027 & 1,970 \\
  \hline
  OpenCV 3.4.0 & C++ & Computer Vision & 400 & 1,168,758 & 3,761 & 7,442\\
  \hline
  POJ-104~\cite{Mou2016} & C++ & Standard Algorithms & 182,815 & 64,518,837 & 312 & 569 \\
  \hline
  Tensorflow~\cite{Abadi} & C++ & Machine learning & 1,584 & 8,444,443 & 5,786 & 11,482 \\
  \hline
  \multirow{4}{*}{GitHub} & C & \multirow{4}{*}{Various} & 42,880 & 89,356,570 & 927 & 1,794\\
                  & Haskell & & 1,371 & 6,745,312 & 4,705 & 7,518\\
                  & OpenCL & & 5,188 & 10,472,388 & 2,299 & 5,132 \\
                  & Swift & & 1,783 & 205,140 & 134 & 371 \\
  \hline
  \textbf{Total} & --- & --- & \textbf{250,428} & \textbf{222,738,929} & \textbf{153,426,059} & \textbf{294,685,614} \\
  \hline
\end{tabular}
	\vspace{.5em}
	\caption{%
		The ten sources of LLVM-IR used to produce datasets for evaluating data flow analyses. Our corpus comprises six programming languages from functional to imperative, high-level to accelerators. The software covers a broad range of disciplines from compilers and operating systems to traditional benchmarks, machine learning systems, and unclassified code downloaded from popular open source repositories.
	}%
	\label{table:corpus} %
\end{table}

We generated a single \textsc{ProGraML} representation for each of the LLVM-IRs, taking an average of 31 ms per IR. Our corpus of unlabeled graphs totals 153M vertices and 295M edges, with an average of 616 vertices per graph with 472 unique vertex representations, and 1183 edges with a maximum edge position of 355 (a large \texttt{switch} statement found in a Tensorflow compute kernel).

\begin{figure}
	\centering
	\begin{subfigure}{.19\linewidth}
		\includegraphics[width=\linewidth]{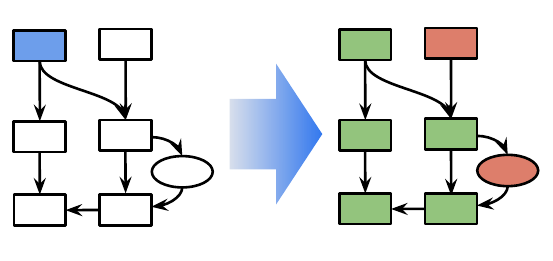}%
		\caption{\textsc{Reachability}.}
		\label{subfig:dataflow_reachability}
	\end{subfigure}
	\hfill
	\begin{subfigure}{.19\linewidth}
		\includegraphics[width=\linewidth]{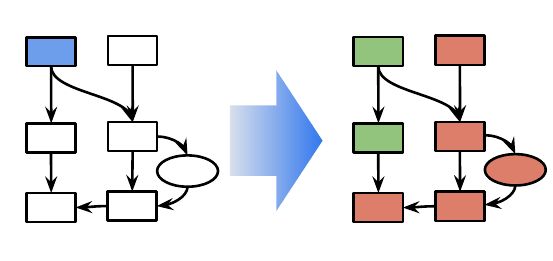}%
		\caption{\textsc{Domtree}.}
		\label{subfig:dataflow_domtree}
	\end{subfigure}
	\hfill
	\begin{subfigure}{.19\linewidth}
		\includegraphics[width=\linewidth]{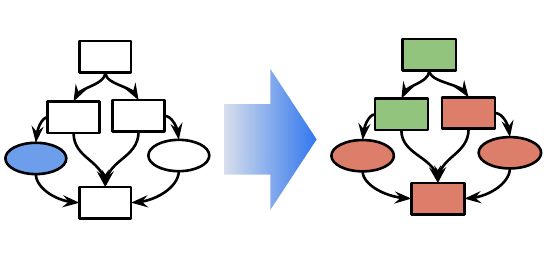}%
		\caption{\textsc{DataDep}.}
		\label{subfig:dataflow_datadep}
	\end{subfigure}
	\hfill
	\begin{subfigure}{.19\linewidth}
		\includegraphics[width=\linewidth]{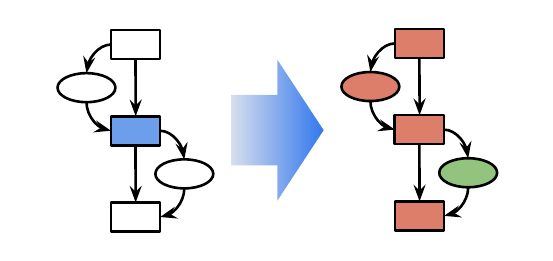}%
		\caption{\textsc{Liveness}.}
		\label{subfig:dataflow_liveness}
	\end{subfigure}
	\hfill
	\begin{subfigure}{.19\linewidth}
		\includegraphics[width=\linewidth]{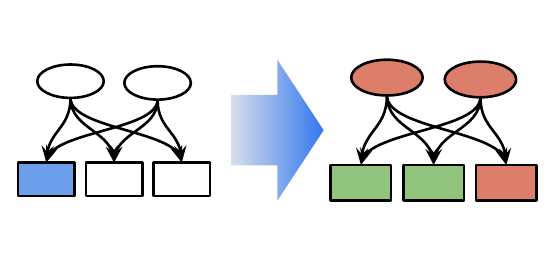}%
		\caption{\textsc{Subexpressions}.}
		\label{subfig:dataflow_subexpressions}
	\end{subfigure}
	\caption{Example input-output graphs for each of the five benchmark compiler analyses. A single vertex is randomly selected from the input graph to represent the starting program for computing the analysis results, indicated using the \emph{vertex selector}. The output graphs contain binary labels for each of the graph vertices after the analysis has completed. As a supervised classification task, the goal of the model is to predict the output vertex labels given the input graph. These small graphs are for illustrative purposes, the LLVM-IR graphs in our real-world corpus contain an average 616 vertices and 1,183 edges.}
	\label{fig:dataflow_examples}%
\end{figure}

We produced labeled graphs from the unlabeled corpus by computing ground truth labels for each of the analysis tasks using a traditional analysis implementation.  For each of the five tasks, and for every unlabeled graph in the corpus, we produce $n$ labeled graphs by selecting unique source vertices $v_{0} \in V$, where $n$ is proportional to the size of the graph:
\begin{equation}
n = \min \left( \left\lceil \frac{|V|}{10} \right\rceil, 10 \right)
\end{equation}
Each instance in the datasets consists of an input graph in which the source vertex is indicated using the \emph{vertex selector}, and an output graph with the ground truth labels used for training or for evaluating the accuracy of model predictions. Figure~\ref{fig:dataflow_examples} illustrates an example input-output instance for each of the five tasks.

We divided the datasets randomly using a 3:1:1 ratio for training, validation, and test instances. The same random allocation of instances was used for each of the five tasks. Where multiple labeled graphs were derived from a single IR, instances derived from the same IR were allocated to the same split.

\subsubsection{Models}%
\label{subsubsec:dataflow_models}

\paragraph{LSTM Baseline} As no prior work offers the expressiveness required to perform the per-statement and per-variable classification required for these analysis tasks, we extended DeepTune~\cite{Cummins2017b}, a state-of-the-art deep learning framework for whole-program classification, to enable per-statement classification. In DeepTune, an OpenCL program is first tokenized and mapped to a sequence of embedding vectors which are then processed through a sequential LSTM model. The final state of the LSTM is optionally concatenated with program-level features, then fed through a fully connected neural network to produce a program-level classification.

\begin{figure}
    \centering %
    \includegraphics[width=.62\columnwidth]{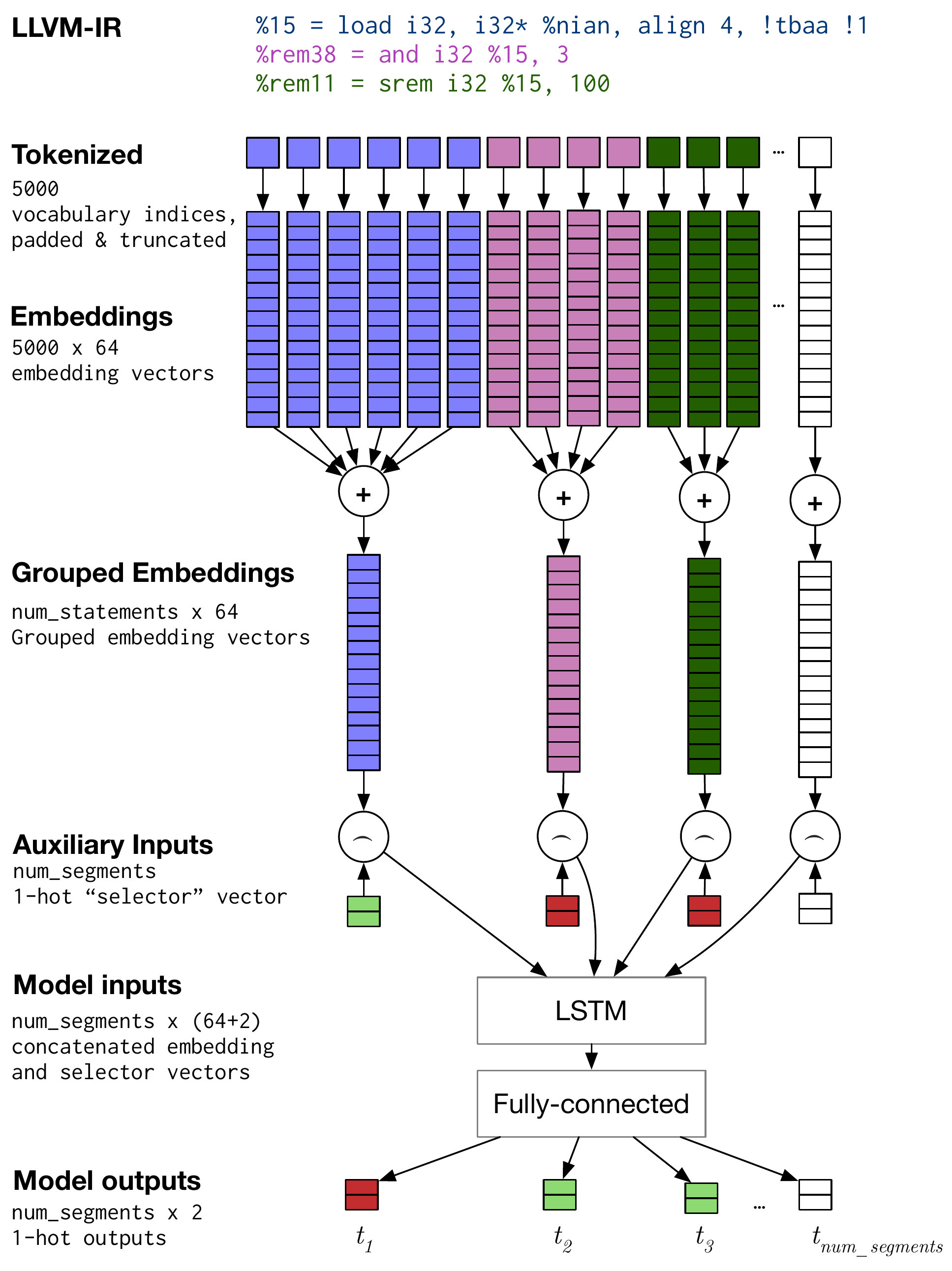}%
    \caption{%
			Extending DeepTune~\cite{Cummins2017b} to perform per-statement classification of an LLVM-IR. In the original work, the latent representation of the entire program token sequence was used for program-level classification, we enable classification of arbitrary token groupings so that we can perform statement-level classification of a program. In the above diagram, $+$ denotes element-wise summation, and $\frown$ denotes vector concatenation.
}%
    \label{figure:lstm_node_level}%
\end{figure}

Figure~\ref{figure:lstm_node_level} shows how we extended this approach for statement-level classification of LLVM-IR. We first replaced the OpenCL tokenizer using one derived from LLVM IR, resulting in a 179-element vocabulary. To adapt the approach for performing statement-level classification, we group embedding vectors by their source statement before using element-wise summation to merge embedding vectors.

We use the same models parameters as in the original work~\cite{Cummins2017b} --- 64-dimensional embedding vectors trained jointly, with 64 sequences per batch, padded and truncated to the same length. As LLVM IR is more verbose than OpenCL, the sequences are significantly longer requiring greater device memory during training and inference. This is a common issue with recurrent neural networks, as sufficient memory is required to store both the intermediate results of the forward pass and the gradients during training~\cite{Ben-Nun2019a}. We found that a sequence length of 5,000 was the maximum that our experimental platform allowed. Where sequences exceed this length, they are truncated, and the model outputs padded with zeros to match the expected shape.

\paragraph{ProGraML} We use the model design outlined in Section~\ref{sec:graph-based-machine-learning} for each of the compiler analysis tasks. While we use the vocabulary of inst2vec, we do not use the pre-trained embeddings, instead initializing the embeddings randomly and training jointly.

Message Passing Neural Networks typically use a small number of propagation steps out of practical consideration for time and space efficiency~\cite{Gilmer2017,Li2015a}, and address problems on smaller graphs than used in this work~\cite{Allamanis2017b}. For a large class of monotone data flow analysis problems, however, it is known that up to $d(G) + 3$ passes over the graph are required, where $d(G)$ is the \emph{loop connectedness} of $G$~\cite{Cooper2003,Kam1977}.
The \emph{loop connectedness} captures the notion of loop-nesting depth in a program and is therefore a program-dependent, but generally unbounded quantity\footnote{Given any depth-first spanning tree (DFST) of $G$, backward edges are defined as edges in $G$ that connect a node to one of its ancestors in the DFST and $d(G)$ is the maximum number of backwards edges in any acyclic path in $G$.}.

We address this challenge with \textsc{ProGraML} by iterating for a fixed number $T$ of message passing steps for training and inference and excluding from the test set graphs for which a traditional implementation of the analysis task requires greater than $T$ iterations to solve. For the experiments in this work we set $T = 30$, leading to 12.56\% of the graphs in the corpus to be excluded across the five tasks. For fairness, we also excluded these graphs from evaluation of the LSTM baseline.

\subsubsection{Training Details and Parameters}
All models were trained in an end-to-end fashion with the Adam optimizer \cite{Kingma2015} using the default configuration and learning rate $0.001$. We trained the models on a NVIDIA GTX 1080 GPU-equipped machine in increments of 10k graphs, testing on a 20k validation set at each checkpoint. Training terminated after six hours, or if accuracy on the validation set reached 99.99\%. After training completed we selected the checkpoint with the highest accuracy on the validation set to use for testing.

\subsection{Case Study B: Heterogeneous Device Mapping}

We apply our methodology to the challenging domain of heterogeneous device mapping (\textsc{DevMap}). Given an OpenCL kernel and a choice of two devices to run it on (CPU or GPU), the \textsc{DevMap} task is to predict the device which will provide the best performance. We chose this problem as it has received significant prior attention, with previous approaches using both hand-engineered features~\cite{Grewe2013} and sequential models~\cite{Cummins2017b, Ben-nun2018}.

\subsubsection{Datasets}

We used the dataset of~\cite{Cummins2017b}, which provides labeled CPU/GPU instances for 256 OpenCL kernels sourced from seven benchmark suites on two combinations of CPU/GPU pairs. The \emph{AMD} set uses an Intel Core i7-3820 CPU and AMD Tahiti 7970 GPU; the \emph{NVIDIA} set uses an Intel Core i7-3820 CPU and an NVIDIA GTX 970 GPU. Each dataset consists of 680 labeled instances derived from the 256 unique kernels by varying dynamic inputs.

\subsubsection{Models}%
\label{subsubsection:devmap_models}

We compare ProGraML with four approaches: First, with a static baseline that predicts the mode device of the dataset distribution. Second, with DeepTune~\cite{Cummins2017b}, which is a sequential LSTM model at the OpenCL source level. Third, to isolate the impact of transitioning from OpenCL source to LLVM-IR, we evaluate against a new DeepTune$_{\text{IR}}$ model, which adapts DeepTune to using tokenized sequences of LLVM-IR as input instead of OpenCL tokens, using the vocabulary described in Section~\ref{subsubsec:dataflow_models}. Finally, we compare against the state-of-the-art approach NCC~\cite{Ben-nun2018}, which replaces the OpenCL tokenizer with a sequence of 200-dimensional embeddings, pre-trained on a large corpus of LLVM-IR using a skip-gram model.

\subsubsection{Training Details and Parameters}

All neural networks are regularized with Dropout \cite{Hinton2012} for generalization and Batch Normalization \cite{Ioffe2015a} in order to be uniformly applicable to vastly different scales of auxiliary input features. We used $10$-fold cross-validation with rotating 80/10/10 splits by training on 80\% of the data and selecting the model with the highest validation accuracy, setting aside $1/10$th of the training data to use for validation. We trained each model for 100 epochs and selected the epoch with the greatest validation accuracy for testing.

\subsection{Case Study C: Algorithm Classification}
In a third case study, we apply our approach to task of classifying algorithms. We use the POJ-104~\cite{Mou2016} dataset.
It contains implementations of 104 different algorithms that were submitted to a judge system. All programs were written by students in higher education. The dataset has around 500 samples per algorithm. We compile them with different combinations of optimization flags to generate a dataset of overall 240k samples. Approximately 10,000 files are held out each as a development and test set.

\subsubsection{Models}%
We compare with recently published tree-based convolutional neural networks (TBCNN)~\cite{Mou2016} and NCC~\cite{Ben-nun2018}, which uses the same approach approach as described in Section~\ref{subsubsection:devmap_models} on this dataset. To further test the expressive power of the graph-based representation against the tree-based (TBCNN) and sequential (NCC) prior work, we present additional experiments: Graph-based baselines based on XFG~\cite{Ben-nun2018} and a \textsc{ProGraML} \emph{structure-only} baseline.

To better understand the qualitative aspects of replacing a graph-based representation that captures program semantics like Contextual Flow Graphs~\cite{Ben-nun2018} (XFG) with the more complete \textsc{ProGraML} representation, we adapted a GGNN~\cite{Li2015a} to directly predict algorithm classes from XFG representations of the programs. In contrast to this, Ben-Nun et al.~\cite{Ben-nun2018} used XFG only to generate statement contexts for use in skip-gram pre-training. Here, we lift this graphical representation and make it accessible to a deep neural network directly, as opposed to the structureless sequential approach in the original work (NCC).

Additionally, we include a \emph{structure-only baseline} of our \textsc{ProGraML} approach, where only the type of each node (instruction, variable, or constant) is encoded, refraining completely from tokenizing statements. We think that algorithm classification is a problem that lends itself especially well to judging the power of the representation \emph{structure}, since most algorithms are well-defined independent of implementation details such as datatypes.

To test the limits of the expressivity of \textsc{ProGraML}, combine our representation with a powerful 10-layer Transformer~\cite{Vaswani2017} encoder model, adapted as a graph neural network for attributed graphs. We induce graph structure by masking the attention scores in the self-attention layer with the adjacency matrix of the \textsc{ProGraML} graphs. A new space-efficient sparse implementation allows processing of graphs with on the order of $10^5$ vertices. Different edge types are encoded by introducing separate \emph{key} and \emph{value} projection matrices into the self-attention layer~\cite{Vaswani2017,Fisches2020}.

\subsubsection{Training Details and Parameters}
The GGNN models were trained with the AdamW~\cite{Loshchilov2019} optimizer with learning rate $2.5\cdot 10^{-4}, \beta_1=0.9, \beta_2=0.999, \varepsilon=10^{-8}$ for 80 epochs. Dropout regularization is employed on the graph states with a rate of $0.2$. The Transformer model uses the same hyperparameters as the GGNN. Additionally a batch size of 64, Dropout regularization of 0.2 on the graph states and weight-sharing between adjacent pairs of layers is employed. The model dimension is equal to the embedding size (200) and the hidden size of the feed-forward layers is 512. The self-attention layers use 5 heads~\cite{Vaswani2017}. Overall, the Transformer model has 5.6 million trainable parameters, around 1.7 million of which are in the embedding layer.
\section{Experimental Results}

This section evaluates the performance and limitations of our approach for the three case studies described in Section~\ref{sec:methodology}, and provides a comparison to state-of-the-art approaches. First we show that \textsc{ProGraML}, unlike prior state-of-the-art approaches to machine learning over code, is capable of replicating core compiler analysis tasks that are key to optimization. Second, we improve upon prior approaches to the task of heterogeneous device mapping. Finally, we set a new state of the art in the difficult task of classifying program algorithms, and ablate the contribution of the structure and content of the \textsc{ProGraML} representation.

\subsection{Case Study A: Compiler Analysis}

Table~\ref{table:data_flow_results} summarizes the performance of the \textsc{ProGraML} approach when tasked with learning a suite of benchmark compiler analysis, along with the performance of a state-of-the-art sequential approach, DeepTune$_\text{IR}$. As a binary classification task, compiler analyses display an extreme class imbalance as only a small fraction of a program graph is typically relevant to computing the result set of an analysis. On our datasets, an accuracy of 96.6\% can be achieved by always predicting the negative class. For this reason we report only binary precision, recall, and $F_1$ scores with respect to the positive class. 

\begin{table}
	\centering%
	\footnotesize
\renewcommand{\arraystretch}{1.6}
\begin{tabular}{L{2.45cm} L{3.1cm} L{3.2cm} L{1.6cm} | R{1.3cm} R{1cm} R{1cm}}
	\toprule
  \textbf{Problem} & \textbf{Analysis type} & \textbf{Example optimization} & \textbf{Model} & \textbf{Precision} & \textbf{Recall} & $\bm{F_1}$\\
  \hline
  \multirow{2}{2.45cm}{\textsc{Reachability}} &
    \multirow{2}{3.1cm}{Forwards, control flow only} &
    \multirow{2}{3.2cm}{Dead code elimination}
        & DeepTune$_{\text{IR}}$ & 0.520 & 0.497 & 0.504\\
                  & & & ProGraML & \textbf{0.997} & \textbf{0.995} & \textbf{0.996}\\
  \hline
  \multirow{2}{2.45cm}{\textsc{DomTree}} &
    \multirow{2}{3.1cm}{Backwards, control flow only} &
    \multirow{2}{3.2cm}{Global Code Motion}
        & DeepTune$_{\text{IR}}$ & 0.721 & 0.081 & 0.138\\
                  & & & ProGraML & \textbf{0.985} & \textbf{0.693} & \textbf{0.781}\\
  \hline
  \multirow{2}{2.45cm}{\textsc{DataDep}} &
    \multirow{2}{3.1cm}{Backwards, control and data flow} &
    \multirow{2}{3.2cm}{Instruction scheduling}
        & DeepTune$_{\text{IR}}$ & 0.999 & 0.136 & 0.236\\
                  & & & ProGraML & \textbf{1.000} & \textbf{0.988} & \textbf{0.993}\\
  \hline
  \multirow{2}{2.45cm}{\textsc{Liveness}} &
    \multirow{2}{3.1cm}{Backwards, control and data flow} &
    \multirow{2}{3.2cm}{Register allocation}
        & DeepTune$_{\text{IR}}$ & --- & --- & ---\\
                  & & & ProGraML & \textbf{1.000} & \textbf{0.999} & \textbf{0.999}\\
  \hline
  \multirow{2}{2.45cm}{\textsc{Subexpressions}} &
  \multirow{2}{3.1cm}{Forwards, statement and operand values and positions} &
  \multirow{2}{3.2cm}{Global Common Subexpression Elimination}
      & DeepTune$_{\text{IR}}$ & \textbf{1.000} & 0.123 & 0.214 \\
                & & & ProGraML & 0.965 & \textbf{0.925} & \textbf{0.930} \\
  \hline
  \multirow{2}{2.45cm}{\textbf{Average}} &
  \multirow{2}{3.1cm}{---} &
  \multirow{2}{3.2cm}{---}
      & DeepTune$_{\text{IR}}$ & 0.810 & 0.209 & 0.273\\
                & & & ProGraML & \textbf{0.989} & \textbf{0.920} & \textbf{0.940}\\
  \hline
\end{tabular}
	\vspace{.7em}
	\caption{%
		Benchmark compiler analyses results using two approaches: (a) DeepTune$_{\text{IR}}$, a sequential model adapted to work at the LLVM-IR level for statement-level classification, and (b) \textsc{ProGraML}, our approach. The relational representation significantly outperforms a sequential approach across all problems. For the Global Common Subexpressions analysis, DeepTune$_{\text{IR}}$ achieved a higher precision than \textsc{ProGraML} by predicting only the root statement as a component in a subexpression, avoiding false-positives.\\*
	}%
	\label{table:data_flow_results} %
\end{table}

As can be seen in Table~\ref{table:data_flow_results}, the relational representation of our approach, coupled with learning through iterative message passing, yields models that far outperform the state-of-the-art sequential approaches to modeling code.  The grouping of program tokens by statement performed by DeepTune$_\text{IR}$ offers a more restrictive classification interface than with \textsc{ProGraML} (where data elements are also represented as graph vertices). As such, it is not able to perform the per-variable classification required for \textsc{Liveness} analysis\footnote{Theoretically the same approach for grouping embeddings by statement could also be extended to group all statements by variables, but this would require duplicating many statements in the input representation, increasing the length of the sequences far beyond what we can practically process using a sequential model.}. For fairness, we exclude \textsc{Liveness} from LSTM aggregate values in Table~\ref{table:data_flow_results}. Despite this, \textsc{ProGraML} achieves an average 94.0 $F_1$, versus 27.3 $F_1$ for DeepTune$_\text{IR}$.

\begin{figure}
	\begin{subfigure}{\linewidth}
		\includegraphics[width=\linewidth]{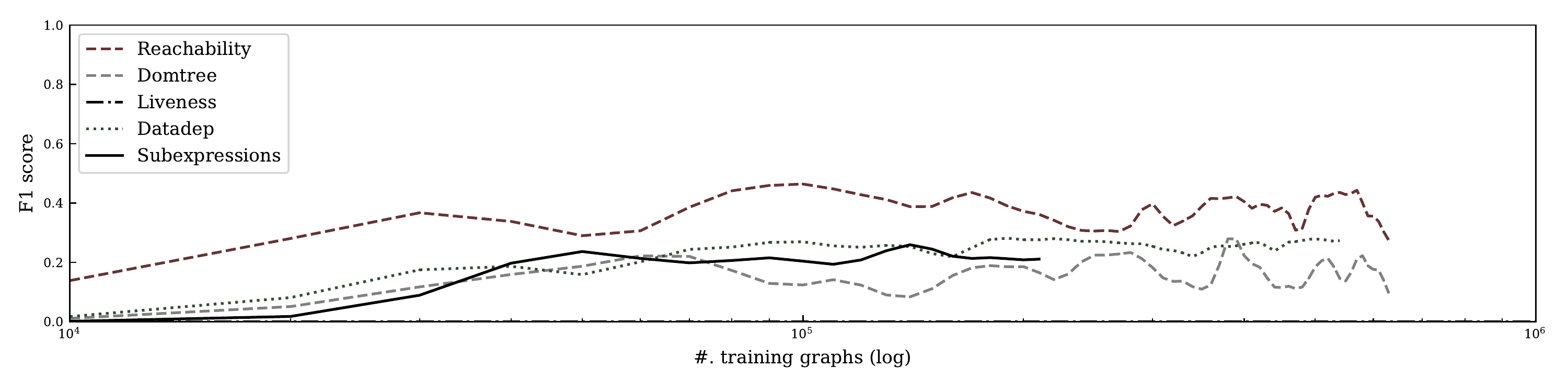}
		\caption{DeepTune$_{\text{IR}}$}
		\label{fig:dataflow-lstm-f1}
	\end{subfigure}
	\\*
	\begin{subfigure}{\linewidth}
		\includegraphics[width=\linewidth]{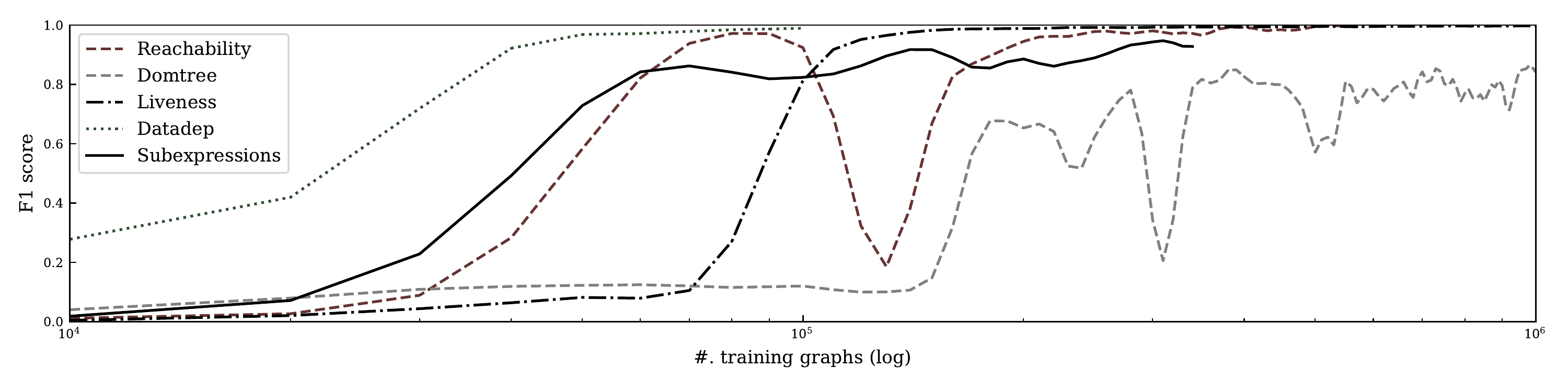}
		\caption{\textsc{ProGraML}}
		\label{fig:dataflow-ggnn-f1}
	\end{subfigure}
	\caption{%
		The $F_1$ score of compiler analysis models on a 20k-graph validation set as a function of the number of training graphs from 10k to 1M. Each model was given six hours to train on a machine with a GTX 1080 GPU, with early termination if accuracy on the validation set reached 99.99\%. We have applied a Gaussian filter ($\sigma=1$) to aid in visualizing the trends in each set of scores.
	}
	\label{fig:dataflow_convergence}
\end{figure}

In many cases, the sequential model regresses to a classification mode in which only the source vertex is labeled as positive, yielding poor recall.  The \textsc{ProGraML} models achieve both high precision and high recall in all tasks except \textsc{DomTree}, where recall is comparatively poor. When model performance is considered as a function of the number of training graphs, as shown in Figure~\ref{fig:dataflow_convergence}, we see that the performance of the \textsc{ProGraML} models quickly convergences towards near-perfect $F_1$ score on a holdout validation set, except in the case of \textsc{DomTree}, where the model is still improving at the end of training. This suggests estimating the \emph{transfer} and \emph{meet} operators of this backwards analysis poses a greater challenge for the network, which may require further training.

Table~\ref{tab:confusion_matrices} shows confusion matrices for the per-vertex classification of \textsc{ProGraML} models on the test set. While the distribution of errors is balanced for \textsc{Reachability}, in the case of \textsc{Domtree} and \textsc{Subexpressions} the ratio of false negatives ($T_+P_-$) outweighs the false positives ($T_-P_+$), indicating that models favor under-approximating the value sets of these analyses. This may be an artifact of training with such a large imbalance towards negatives ($T_-$) over positive ($T_+$) class labels. In future work we will explore addressing this class imbalance by selecting multiple root points on a graph for analysis simultaneously, thereby increasing the size of the value set to include multiple (possibly overlapping) regions.

\begin{table}
	\centering
\scriptsize
\renewcommand{\arraystretch}{1.5}
\begin{subfigure}{.19\linewidth}
	\centering
	\begin{tabular}{r | c c}
		\toprule
		& $\bm{P_-}$ & $\bm{P_+}$ \\
		\midrule
		$\bm{T_-}$ & 96.64\% & 0.01\% \\
		$\bm{T_+}$ & 0.01\% & 3.34\% \\
		\bottomrule
	\end{tabular}
	\caption{\textsc{Reachability}}
\end{subfigure}
\hfill
\begin{subfigure}{.19\linewidth}
	\centering
	\begin{tabular}{r | c c}
		\toprule
		& $\bm{P_-}$ & $\bm{P_+}$ \\
		\midrule
		$\bm{T_-}$ & 95.94\% & 0.13\% \\
		$\bm{T_+}$ & 1.70\% & 2.23\% \\
		\bottomrule
	\end{tabular}
	\caption{\textsc{DomTree}}
\end{subfigure}
\hfill
\begin{subfigure}{.19\linewidth}
	\centering
	\begin{tabular}{r | c c}
		\toprule
		& $\bm{P_-}$ & $\bm{P_+}$ \\
		\midrule
		$\bm{T_-}$ & 98.80\% & 0.00\% \\
		$\bm{T_+}$ & 0.00\% & 1.20 \% \\
		\bottomrule
	\end{tabular}
	\caption{\textsc{DataDep}}
\end{subfigure}
\hfill
\begin{subfigure}{.19\linewidth}
	\centering
	\begin{tabular}{r | c c}
		\toprule
		& $\bm{P_-}$ & $\bm{P_+}$ \\
		\midrule
		$\bm{T_-}$ & 90.00\% & 0.00\% \\
		$\bm{T_+}$ & 0.00\% & 9.99\% \\
		\bottomrule
	\end{tabular}
	\caption{\textsc{Liveness}}
\end{subfigure}
\hfill
\begin{subfigure}{.19\linewidth}
	\centering
	\begin{tabular}{r | c c}
		\toprule
		& $\bm{P_-}$ & $\bm{P_+}$ \\
		\midrule
		$\bm{T_-}$ & 99.12\% & 0.02\% \\
		$\bm{T_+}$ & 0.13\% & 0.73\% \\
		\bottomrule
	\end{tabular}
	\caption{\textsc{Subexpressions}}
\end{subfigure}
	\caption{%
		Confusion matrices for compiler analyses using \textsc{ProGraML}. Rows denote true negative ($T_-$) and true positive ($T_+$), columns denote predicted negative ($P_-$) and predicted positive ($P_+$). The value of a cell is the ratio of per-vertex model outputs of this type, e.g. $T_-P_+$ is the ratio of false positives.%
	}
	\label{tab:confusion_matrices}
\end{table}

\subsection{Case Study B: Heterogeneous Device Mapping}

The performance of ProGraML and baseline models is shown in Table~\ref{figure:devmap_results}. We reused the pre-trained inst2vec-embeddings for the NCC baseline that were published with the original work, however all models themselves have been reimplemented to ensure fair comparison across different models under a unified evaluation regime and absolute performance numbers can thus deviate from the original publications.

Baseline models were trained with hyperparameters from the original works. For the \textsc{ProGraML} results we used 6 layers in the GGNN corresponding to 6 timesteps of message propagation, while sharing parameters between even and odd layers to introduce additional regularization of the weights. We ran a sweep of basic hyperparameters which led us to use the pre-trained inst2vec statement embeddings~\cite{Ben-nun2018} and to exclude the use of position representations. Both of these hyperparameter choices help generalization by reducing the complexity of the model. This is not surprising in light of the fact that the dataset only contains 680 samples derived from 256 unique programs. \textsc{ProGraML} was trained with the Adam optimizer with default parameters, a learning rate of $10^{-3}$ and a batch size of 18,000 nodes for 300 epochs (resulting in ca. 12000 iteration steps of the optimizer). Additionally we found dropout~\cite{Srivastava2014} with a rate of $0.1$ on the weights of the message propagation function to be beneficial on the validation set as well. For the \textsc{ProGraML} result, we repeat the automated sweep for all hyperparameter configurations and picked the configuration with the best average validation performance. Performance on the unseen tenth of the data is reported.

As can be seen, \textsc{ProGraML} outperforms prior approaches to this problem by all metrics (accuracy, precision, recall, and $F_1$), across both device datasets.

\begin{table}
	\centering%
	\centering
\footnotesize
\renewcommand{\arraystretch}{1.6}
\begin{subfigure}{.48\linewidth}
\begin{tabular}{l llll}
	\toprule
	 & Accuracy & Precision & Recall & $F_1$\\
	\midrule
	Static Mapping                 & 58.8\% & 0.35 & 0.59 & 0.44\\
	DeepTune~\cite{Cummins2017b}   & 71.9\% & 0.72 & 0.72 & 0.72\\
	DeepTune$_{\text{IR}}$         & 73.8\% & 0.76 & 0.74 & 0.75\\
	NCC~\cite{Ben-nun2018}    & 80.3\% & 0.81 & 0.80 & 0.80\\
	ProGraML                       & \textbf{86.6}\% & \textbf{0.89} & \textbf{0.87} & \textbf{0.88}\\
	\midrule
\end{tabular}
\caption{AMD}
\end{subfigure}
\begin{subfigure}{.48\linewidth}
\begin{tabular}{l llll}
	\toprule
	 & Accuracy & Precision & Recall & $F_1$\\
	\midrule
	Static Mapping                 & 56.9\%    & 0.32 & 0.57 & 0.41\\
	DeepTune~\cite{Cummins2017b}   & 61.0\%    & 0.69 & 0.61 & 0.65\\
	DeepTune$_{\text{IR}}$         & 68.4\% & 0.70 & 0.68 & 0.69\\
	NCC~\cite{Ben-nun2018}    & 78.5\%   & 0.79 & 0.79 & 0.79\\
	ProGraML                       & \textbf{80.0}\% & \textbf{0.81} & \textbf{0.80} & \textbf{0.80}\\
	\midrule
\end{tabular}
\caption{NVIDIA}
\end{subfigure}
	\caption{%
		Five approaches to predicting heterogeneous device mapping:
		(a) Static Mapping
		(b) DeepTune~\cite{Cummins2017b}, a sequential model using tokenized OpenCL,
		(c) DeepTune$_{\text{IR}}$, the same model adapted for tokenized LLVM-IR,
		(d) NCC, which uses pre-trained statement embeddings, and
		(e) \textsc{ProGraML}, our approach.%
	}%
	\label{figure:devmap_results} %
\end{table}

\subsection{Case Study C: Algorithm Classification}
Table \ref{tab:classify} summarizes the algorithm classification accuracy results of our method and baselines.

\paragraph{Baseline Experiments} Lifting the XFG graph representation from pretraining embeddings~\cite{Ben-nun2018} to the high-level task of algorithm classification showed a strong improvement of performance. We found that jointly learning the embeddings with the model from scratch (cf. \emph{XFG-rnd} in Table \ref{tab:classify}) outperformed both fixed inst2vec embeddings (\emph{XFG-i2v}) as well as finetuning of inst2vec embeddings (not shown). Both baselines are an improvement over inst2vec and show that graph-based models are more suited to the task than models with less structure.

Next, we want to understand whether our particular graph-representation reached its design goals and can provide additional improvement on POJ104.

\paragraph{\textsc{ProGraML} Experiments}
To ablate the contribution of the tokenization from the performance boost provided by the \textsc{ProGraML} representation itself, we include a GGNN-based \emph{structure-only baseline} (denoted as \emph{GGNN-s}) of our approach, where the only information on each node is whether it represents a statement or an identifier in the graph, but we refrain from tokenizing statements.
We think that algorithm classification is a problem that lends itself especially well to judging the power of the representation \emph{structure}, since most algorithms are well-defined independent of implementation details, such as datatypes.

The results in Table \ref{tab:classify} show that the structure alone of our \textsc{ProGraML} representation is sufficient to outperform the XFG baselines and prior work on the task of algorithms classification. Adding inst2vec statement tokenization further improves performance. This suggests that there is room for improvement in performance by extending the graph encoding method to achieve better vocabulary coverage and stronger generalization. We leave this to future work.

\begin{table}
	\centering
	\footnotesize
	\begin{tabular}{lllllllll}
		\toprule
			  & TBCNN~\cite{Mou2016} & NCC~\cite{Ben-nun2018} & \multicolumn{2}{c}{XFG}  & \multicolumn{3}{c}{ProGraML}\\
		\cmidrule(lr){4-5} \cmidrule(lr){6-8}
		Metric & & & i2v & rnd  & GGNN-s & GGNN & Transformer\\
		\midrule
		Test Error [\%] & 6.0 & 5.17 & 4.56 & 4.29 & 3.87 & 3.78 & \textbf{3.33}\\ 
		Improvement [\%] & --- & 0.0 & 11.8 & 17.0 & 25.1 & 26.9 & \textbf{35.6}\\
		\bottomrule
		\vspace{.7em}
	\end{tabular}
		\caption{Algorithm Classification Error on POJ-104 \cite{Mou2016}. The two XFG models are distinct only in their embedding layers: \emph{XFG-i2v} uses inst2vec embeddings, while \emph{XFG-rnd} jointly learns the embeddings. The results denoted by Surface Features and TBCNN are reproduced from Mou et al. \cite{Mou2016}.}
\label{tab:classify}
\end{table}

\section{Related Work}

In order to perform machine learning on programs, prior work has employed methods from Natural Language Processing and represented programs as a sequence of lexical tokens~\cite{Allamanis2013a, Allamanis2016d, Cummins2017b}. However, it has been observed~\cite{Raychev2015,Allamanis2017b,Alon2018c} that it is critical to capture the structured nature of programs and syntactic (tree-based) as well as semantic (graph-based) representations have been proposed~\cite{Allamanis2017a,Brauckmann2020}.
There is a line of research that considers program representations based on Abstract Syntax Trees (ASTs):
Dam et al.~\cite{Dam2018} annotate nodes in the AST with type information and employ Tree-Based LSTMs~\cite{Tai2015a} for program defect prediction.
Both Raychev et al.~\cite{Raychev2015} and Alon et al.~\cite{Alon2018a,Alon2018c} use path-based abstractions of the AST as program representations, while Allamanis et al.~\cite{Allamanis2017b} augment ASTs with a hand-crafted set of additional typed edges and use GGNNs~\cite{Li2015a} to learn downstream tasks related to variable naming. Another line of research considers modelling binary similarity via control-flow graphs (CFGs) with an adaptation of GNNs called Graph Matching Networks~\cite{Li2019}.

The history of representing programs as graphs for optimization goes back to Program Dependence Graphs (PDGs)~\cite{Ferrante1987}, which remove superfluous control flow edges to ease optimization with a compact graph representation. A more contemporary precursor of our \textsc{ProGraML} representation ConteXtual Flow Graphs (XFGs)~\cite{Ben-nun2018}, which combine control flow with dataflow in order to learn unsupervised embeddings of LLVM-IR statements. While \textsc{ProGraML} is designed to extend the concepts of XFGs, \textsc{ProGraML} preserve the notion of argument order and includes nodes for both variables and constant values and all control flow edges. These changes reflects the design goals of the representations --- XFGs are designed to easily express program semantics by omitting superfluous control relations and other execution-specific properties. \textsc{ProGraML}, in combining CG, CFG, and DFG, offers a compiler-level program representation that is designed to be useful for a variety of purposes from specific program analyses to downstream optimization tasks.
 
Another approach is taken by IR2Vec~\cite{KeerthyS2019}, which defines an LLVM-IR-specific statement representation that elegantly models part-of-statements as relations. However, in order to compute the values of the embeddings, IR2Vec requires access to the type of data flow analyses that our approach is learning from data alone.

Brauckmann et al.~independently proposed a graph-based representation based on Control and Data Flow Graphs (CDFG)~\cite{Brauckmann2020}. As in this work, CDFGs represent instructions as graph vertices and have bi-directional edges for control and data relations. What differentiates \textsc{ProGraML} from CDFGs is the richness of the program representation: CDFGs ignore the data elements of programs, and only  instruction opcodes are used for vertex embeddings. The latent representation of statements is thus invariant to instruction operands, their order, data types, and instruction modifiers. This limits the representational power of the approach, e.g. by omitting code properties required for reasoning about variables and expressions, as we explore through the \textsc{Liveness} and \textsc{Subexpressions} experiments in this work. Finally, our approach provides $76.6\times$ improved inference throughput, though we suspect this may be an artifact of our implementation as both approaches scale linearly w.r.t.\ the size of the graph and number of message passing steps.

Graph Neural Networks comprise a diverse class of neural networks that learn by propagating information along edges~\cite{Gori2005a,Scarselli2009,Battaglia2018}. Approaches based on Recurrent Neural Networks (RNNs)~\cite{Li2015a,Gilmer2017} as well as convolutional~\cite{Defferrard2016,Kipf2017} and attention-based methods~\cite{Velickovic2018,Wang2019a} exist. GNNs as a family have been shown to have enough expressive power to address difficult combinatorial problems like the graph isomorphism test~\cite{Xu2019} at least as well as the Weisfeiler-Lehman algorithm~\cite{Weisfeiler1968}. Please refer to Battaglia et al.~\cite{Battaglia2018} for a comprehensive review of GNNs.

\section{Conclusions}

With the demand for aggressively optimizing compilers increasing, there is an increasing burden on compiler heuristics to make good optimization decisions. While tuning heuristics by hand is expensive and slow to keep up with the pace of compiler and architecture advancements, machine learning offers tremendous benefits for automatically constructing heuristics that are both cheaper to develop and better performing than hand-crafted equivalents. The success of these machine learning approaches is bound by the quality of the input used to represent programs, and the ability of models to process these representations.

In this work, we present a graph-based representation for programs, derived from compiler IRs, that accurately captures the semantics of a program's statements and the relations between them. Our approach is more expressive than prior sequence- or graph-based representations, while closely approximating the representations that are traditionally used within compilers.

We have shown through a constructivist approach that machine learning is capable of approximating the types of compiler analyses that are key for optimizations. In testing our approach on a suite of established compiler tasks that even state-of-the-art machine learning methods struggle with, our goal is to inspire confidence in machine learning as a viable tool for reasoning about program semantics, as opposed to a black box which discourages, rather than inspires, a more systematic approach to reasoning about optimizations. When tasked with real-world problems spanning multiple domains and source languages, our approach outperforms prior state-of-the-art approaches.

Our hope in developing \textsc{ProGraML} is to provide a re-usable toolbox for representing and reasoning about programs that can be used for a wide variety of downstream tasks. Promising research avenues for downstream tasks enabled by our enriched program representation and the ability to perform statement-level inference include automatic parallelization, static performance estimation, and IR-to-IR transpilation. Additionally, while the applications of deep learning to compilers is rapidly evolving~\cite{Allamanis2017a,Cummins2020}, we hope to focus attention on the challenges that machine learning methods face in the domain of programming languages: scalability when faced with large inputs, modeling very-long-range dependencies, and learning over unbounded vocabularies.

\bibliographystyle{unsrt}
\bibliography{refs}

\end{document}